\documentclass{template}

\newcommand{\templateoption}{option1}

\usepackage[T1]{fontenc}
\usepackage{lmodern}  
\usepackage{xspace}
\usepackage{soul, algorithm, algpseudocode, subcaption, minted, stfloats, xcolor, relsize, booktabs, multirow, makecell, textpos, url}
\usepackage{graphicx,amsmath,amssymb,hyperref}
\usepackage[authoryear,round]{natbib}
\usepackage{xcolor}
\usepackage{ifthen}

\usepackage{siunitx}
\newlength{\savewidth}
\newcommand\shline{\noalign{\global\savewidth\arrayrulewidth
  \global\arrayrulewidth 1pt}\hline
  \noalign{\global\arrayrulewidth\savewidth}}

\hypersetup{
  colorlinks=true,
  urlcolor=magenta,
  linkcolor=black,
  citecolor=link,
  filecolor=black,
  pdfborder={0 0 0}
}

\definecolor{link}{HTML}{0063BE}

\providecommand{\sectionname}{Section}
\newcommand{\secref}[1]{%
  \sectionname~\hyperref[#1]{\textcolor{link}{\ref*{#1}}}%
}
\providecommand{\equationname}{Equation}
\newcommand{\eqrefc}[1]{%
  \equationname~\hyperref[#1]{\textcolor{link}{\ref*{#1}}}%
}
\newcommand{\figref}[1]{%
  \figurename~\hyperref[#1]{\textcolor{link}{\ref*{#1}}}%
}
\newcommand{\subfigref}[2]{%
  \figurename~\hyperref[#1]{\textcolor{link}{\ref*{#1}#2}}%
}
\newcommand{\tabref}[1]{%
  \tablename~\hyperref[#1]{\textcolor{link}{\ref*{#1}}}%
}
\newcommand{\appendixref}[1]{%
  Appendix~\hyperref[#1]{\textcolor{link}{\ref*{#1}}}%
}
\newcommand{\tablestyle}[2]{\setlength{\tabcolsep}{#1}\renewcommand{\arraystretch}{#2}\centering\footnotesize}
\renewcommand{\paragraph}[1]{\vspace{1.25mm}\noindent\textbf{#1}}
\newcommand{\methodname}{Derf}

\newcommand{\github}{\raisebox{-1.3pt}{\includegraphics[height=1.05em]{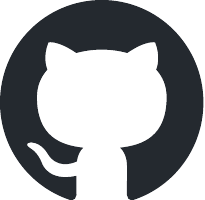}}\xspace}
\newcommand{\worldwideweb}{\raisebox{-1.3pt}{\includegraphics[height=1.05em]{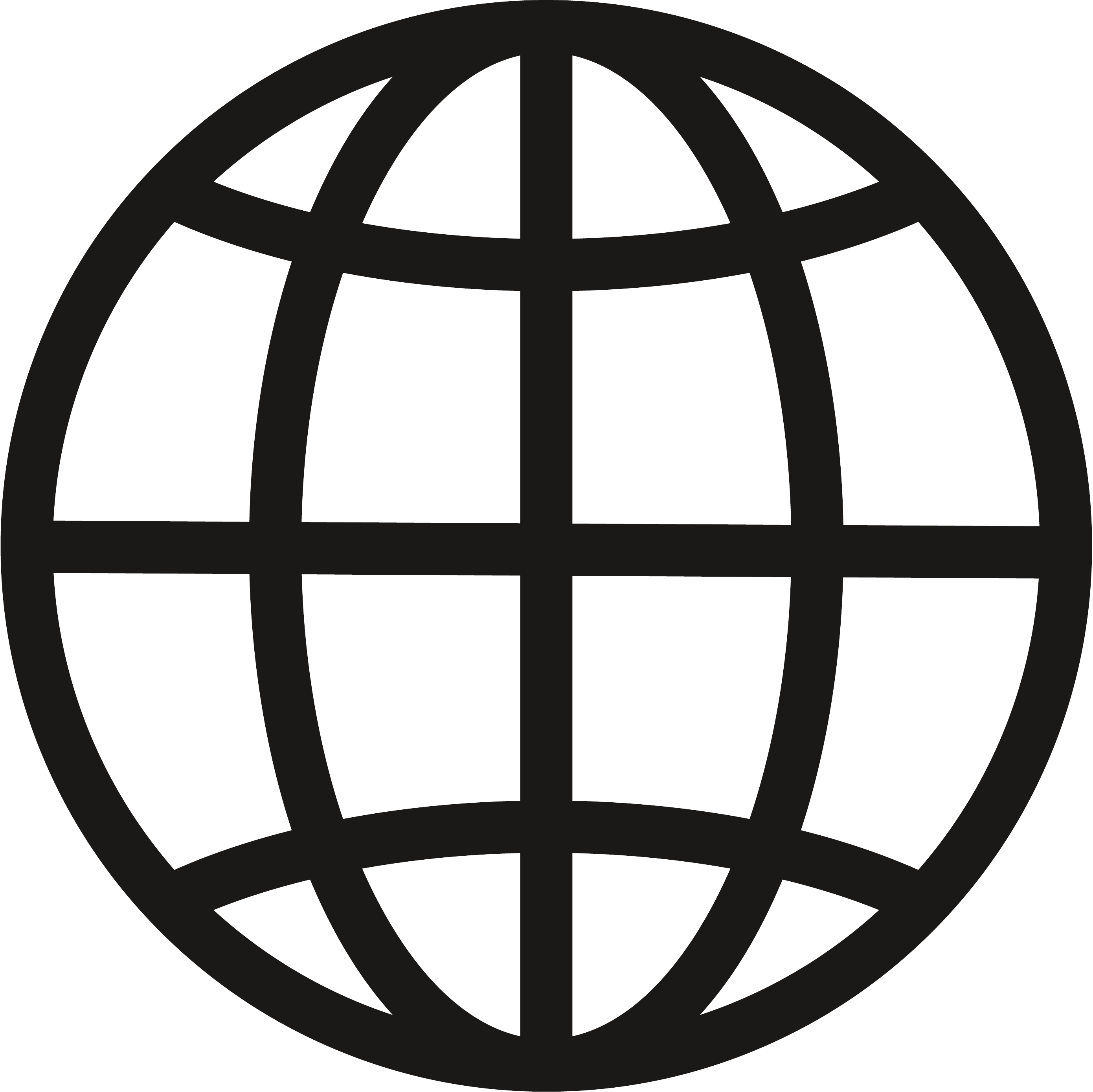}}\xspace}
\newcommand{\hf}{\raisebox{-1.3pt}{\includegraphics[height=1.05em]{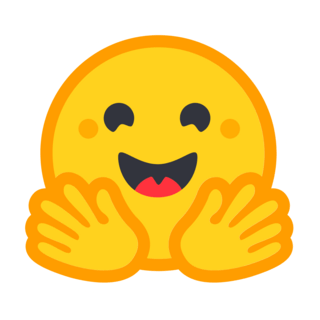}}\xspace}

\makeatletter
\fancypagestyle{firststyle}{
  \fancyhead[L]{}
  \fancyhead[C]{}
  \fancyhead[R]{}
  \fancyfoot[L]{\footerfont \the\correspondingauthor}
  \fancyfoot[R]{}
}
\makeatother

\makeatletter
\DeclareRobustCommand\bfseries{%
  \not@math@alphabet\bfseries\mathbf
  \fontseries\bfdefault\selectfont
  \sffamily
}
\makeatother

\DeclareTextFontCommand{\textbf}{\bfseries\sffamily}

\title{\centering \fontsize{19}{16}\selectfont{Stronger Normalization-Free Transformers}}

\author{
    \vspace{.05cm}
    \parbox{\textwidth}{\centering
        Mingzhi Chen\textsuperscript{1} \quad
        Taiming Lu\textsuperscript{1} \quad
        Jiachen Zhu\textsuperscript{2} \quad
        Mingjie Sun\textsuperscript{3} \quad
        Zhuang Liu\textsuperscript{1} 
    }
    \\
    {\normalfont\fontsize{11}{15}\selectfont {\textsuperscript{1}Princeton University}\hspace{.1cm}}
    {\normalfont\fontsize{11}{15}\selectfont {\textsuperscript{2}New York University}\hspace{.1cm}}
    {\normalfont\fontsize{11}{15}\selectfont {\textsuperscript{3}Carnegie Mellon University}}
}

\makeatletter
\renewcommand{\abscontent}{}%
\makeatother

\newenvironment{abstractblock}{%
  {\centering\large\bfseries\sffamily Abstract\par}
  \vspace{0.3em}
  \begin{list}{}{%
      \setlength{\leftmargin}{2em}
      \setlength{\rightmargin}{2em}
      \setlength{\topsep}{0pt}
      \setlength{\parsep}{0pt}
  }
  \item[]
}{%
  \end{list}
  \par\normalfont\vspace{1em}
}

\newlength{\abstractwidth}
\begin{document}

\begingroup
\makeatletter
\let\raggedright\centering
\makeatother

\maketitle
\endgroup

\newcommand{\abstractcontent}{%
Although normalization layers have long been viewed as indispensable components of deep learning architectures, the recent introduction of Dynamic Tanh (DyT) \citep{zhu2025transformers} has demonstrated that alternatives are possible. The point-wise function DyT constrains extreme values for stable convergence and reaches normalization-level performance; this work seeks further for function designs that can surpass it. We first study how the intrinsic properties of point-wise functions influence training and performance. Building on these findings, we conduct a large-scale search for a more effective function design. Through this exploration, we introduce $\mathrm{Derf}(x) = \mathrm{erf}(\alpha x + s)$, where $\mathrm{erf}(x)$ is the rescaled Gaussian cumulative distribution function, and identify it as the most performant design. \methodname{} outperforms LayerNorm, RMSNorm, and DyT across a wide range of domains, including visual recognition and generation, speech representation, and DNA sequence modeling. Our analysis also suggests that the performance gains of \methodname{} largely stem from its improved generalization rather than stronger fitting capacity. Its simplicity and stronger performance make \methodname{} a practical choice for normalization-free Transformer architectures.  Our code is available at this {\href{https://github.com/zlab-princeton/Derf}{\texttt{link}}.}
}

\newcommand{\teaserfigure}[2]{%
\begin{figure}[#1]
    \centering
    \begin{minipage}{\abstractwidth}
        {\centering

        \vspace{-.5cm}
        
        \hspace{1em} \includegraphics[width=.95\linewidth]{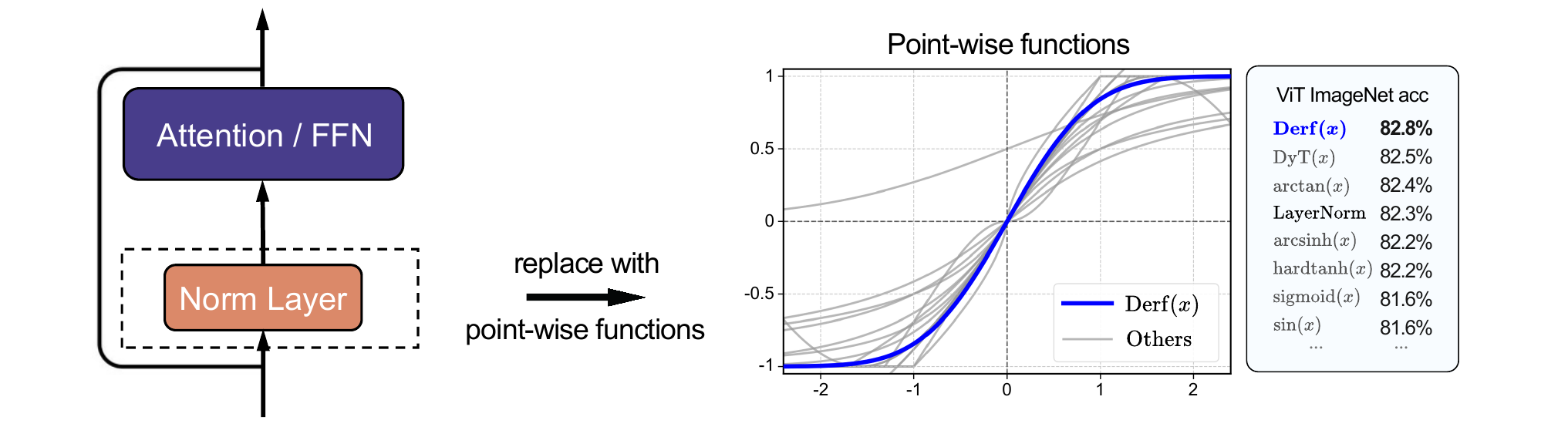}

        \vspace{-0.2em}

        \small \textbf{a)} We search point-wise functions of different shapes as norm layer replacement. \\[0.5em]

        \vspace{1.2em}
        }

        \begin{minipage}[b]{0.58\linewidth}
            \centering
            \includegraphics[width=.9\linewidth]{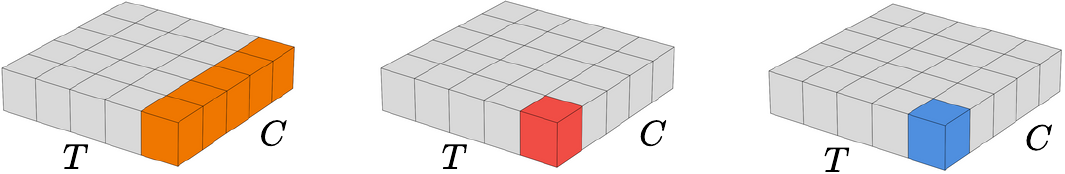}

            \vspace{0.5em}
            {
            \textcolor{orange}{\hspace{0.1cm}{LayerNorm}}:  $\tfrac{x - \mu}{\sqrt{\sigma^2 + \epsilon}}$%
             \hspace{0.3cm}
            \textcolor{red}{DyT}: $\tanh(\alpha x)$%
             \hspace{0.3cm}
            \textcolor{blue}{Derf}: $\mathrm{erf}(\alpha x + s)$
            }

            \vspace{0.5em}
            \small \textbf{b)} Formulation of LayerNorm (LN), DyT, and \methodname{} (ours).
        \end{minipage}
        \hfill
        \begin{minipage}[b]{0.40\linewidth}
            \vspace*{-10em}
            \centering
            \resizebox{.95\linewidth}{!}{
\begin{tabular}{lccc}
\toprule
method & ViT acc ($\uparrow$) & DiT FID ($\downarrow$) & DNA acc ($\uparrow$) \\ 
\midrule 
LN & \(82.3\%\) & \(45.91\) & \(86.9\%\) \\ 
DyT & \(82.5\%\) & \(45.66\) & \(86.9\%\) \\ 
\methodname{} &
  \(\mathbf{82.8\%}\) &
  \(\mathbf{43.94}\) &
  \(\mathbf{87.3\%}\) \\
\midrule
\end{tabular}

            }
        
            \vspace{1.5em}
            \small \textbf{c)} Performance across domains.
        \end{minipage}

        \vspace{.3cm}
        
        \caption{
            \textbf{We introduce Dynamic erf (\methodname{}), a point-wise function to replace normalization layers, which outperforms other normalization-based and -free designs across domains.} 
            (a) We identify the feasible function shape for replacing the normalization layer and propose a large set of point-wise functions within this space. Evaluating all candidates, we identify and introduce \methodname{} as the strongest choice. (b) LayerNorm, DyT \citep{zhu2025transformers}, and \methodname{} operate in fundamentally different ways: with channels $C$ and tokens $T$, LayerNorm normalizes each token across the channel axis, whereas DyT and \methodname{} apply independent scalar mappings to each element. (c) Across ImagenNet-1K classification and generation, and DNA modeling, \methodname{} consistently outperforms LayerNorm and DyT. \methodname{} demonstrates that a point-wise function can not only replace normalization but also surpass it.
        }
        \label{fig:teaser}
    \end{minipage}
\end{figure}%
}

\newcommand{\linkstablestyleone}{%
    \begin{center}
        \small
        \renewcommand{\arraystretch}{1.2}
        \begin{tabular}{rll}
            \worldwideweb & \textbf{Website} & \url{https://www.test.com/}\\
            \github & \textbf{Code} & \url{https://github.com/test}\\
            \hf & \textbf{Data} & \url{https://huggingface.com}
        \end{tabular}
    \end{center}%
}

\newcommand{\linkstablestyletwo}{%
    \noindent\small
    \textbf{Website:} \url{https://www.test.com/}\\
    \textbf{Code:}    \url{https://github.com/test}\\
    \textbf{Data:}    \url{https://huggingface.com}
}


\makeatletter
\ifthenelse{\equal{\templateoption}{option1}}{

    \vspace{-.5cm}
    
    \teaserfigure{!ht}{.8\textwidth}

    \vfill

    \begin{abstractblock}
    \abstractcontent
    \end{abstractblock}

    \newpage
}{
    \ifthenelse{\equal{\templateoption}{option2}}{
        \vspace{0.1cm}
        
        \begin{abstractblock}
        \abstractcontent
        \end{abstractblock}


        \teaserfigure{!bh}{.50\textwidth}

        \newpage
    }{
        \begin{abstractblock}
        \abstractcontent
        \end{abstractblock}

        \vspace{0.5cm}


        \vspace{-0.5cm}

    }
}
\makeatother

\section{Introduction}

\label{sec:intro}

Normalization layers have become a critical component in modern deep neural networks. Since the invention of Batch Normalization \citep{ioffe2015batch}, more and more variants have been developed to adapt normalization to various architectures and model modalities \citep{ba2016layer, salimans2016weight, ulyanov2016instance, wu2018group, zhang2019root}. By regulating the distribution of intermediate activations, normalization layers have long demonstrated their strong capability in stabilizing training and accelerating model convergence \citep{santurkar2018does, bjorck2018understanding}. 

Due to the inherent formulation of normalization layers, they heavily rely on activation statistics during training. This introduces additional memory access and synchronization overhead \citep{zhang2019root, chen2020effective, yang2022unified}. Moreover, some normalization methods are highly sensitive to batch size, and inappropriate batch settings can lead to unstable training \citep{wu2018group, lian2019revisit, singh2020filter}. These issues motivate recent efforts to develop normalization-free methods. Among these attempts, Dynamic Tanh \citep{zhu2025transformers}, an S-shaped point-wise function, has emerged as a simple yet effective drop-in replacement for normalization layers. This work has established the foundation for point-wise functions that match the performance of normalization layers, yet functions that can surpass them remain unexplored. In this work, we aim to discover point-wise functions that outperform normalization layers to push toward stronger Transformer architectures \citep{vaswani2017attention, dosovitskiy2020image}.

We first systematically study how the intrinsic properties of point-wise functions affect the training dynamics and final performance. Specifically, we focus on four fundamental and representative function properties: \textit{zero-centeredness}, \textit{boundedness}, \textit{center sensitivity}, and \textit{monotonicity}. Each property is independently examined through controlled experiments on a diverse set of point-wise functions to assess its impact on the training result. This analysis isolates a subset of point-wise functions as effective normalization replacements and yields a concrete design principle for normalization-free Transformers.

Guided by these principles, we identify a set of promising point-wise functions that have the potential to surpass the performance of normalization layers. Within this set, we empirically search for the optimal designs, among which Dynamic erf (\methodname{}) emerges as a simple yet the most performant function (\subfigref{fig:teaser}{a}). 
\methodname{} augments $\mathrm{erf}(x)$ with learnable parameters, where 
the error function $\mathrm{erf}(x)$ is an S-shaped, rescaled cumulative distribution of a standard Gaussian around zero.

We evaluate \methodname{} spanning multiple modalities (vision, language, speech, and DNA sequences); covering various tasks (classification, generation, and sequence modeling), under different training paradigms (supervised and self-supervised). Across all these settings, \methodname{} consistently surpasses LayerNorm, and Dynamic Tanh (\subfigref{fig:teaser}{b,1c}). To pinpoint the source of these gains, we measure the training loss in evaluation mode after optimization. \methodname{} exhibits higher training loss than normalization-based models, indicating that its superior performance stems from stronger generalization rather than enhanced fitting capacity. Overall, our work demonstrates that well-designed point-wise functions can outperform normalization layers.

\section{Background}
\label{sec:background}

\paragraph{Normalization layers.} Normalization layers have become pivotal components of modern neural networks. Among the various normalization techniques, Batch Normalization (BN) \citep{ioffe2015batch}, Layer Normalization (LN) \citep{ba2016layer}, and Root Mean Square Normalization (RMSNorm) \citep{zhang2019root} are the three most widely used in deep learning models.
\begin{equation}
    y =  \gamma * \frac{x - \mu}{\sqrt{\sigma^2 + \epsilon}} + \beta
    \label{eq:normalization}
\end{equation}
All normalization methods adhere to a unified paradigm, formalized in \eqrefc{eq:normalization}, where activations within each group are centered and scaled by their mean $\mu$ and standard deviation $\sigma$ (with $\epsilon$ for numerical stability) to maintain consistent scale and stable gradient flow.
The main distinction among different normalization methods lies in how the activations are grouped when computing $\mu$ and $\sigma$. 
For example, LN computes the statistics along the channel dimension for each token independently. Given a token representation $x \in \mathbb{R}^C$, the mean and variance are computed as \eqrefc{eq:ln_stats}, 
where $C$ denotes the number of hidden features (channels). Due to its per-token normalization, LN is particularly well-suited for Transformer architectures, where activations across tokens exhibit diverse statistics.
\begin{equation}
    \mu = \frac{1}{C} \sum_{k=1}^{C} x_k, \qquad
    \sigma^2 = \frac{1}{C} \sum_{k=1}^{C} (x_k - \mu)^2,
    \label{eq:ln_stats}
\end{equation}

\paragraph{Point-wise functions.}
The strong reliance of normalization layers on activation statistics has motivated further exploration of statistics-free methods \citep{he2023simplifying, heimersheim2024you, jha2024aero, zhu2025transformers}. Among these approaches, point-wise functions \citep{zhu2025transformers} have emerged as simple yet effective alternatives to traditional normalization methods. Unlike normalization, a point-wise function applies the same parametric mapping $f(x; \theta)$ to each activation independently. The parameters $\theta$ are fixed or learned, rather than being computed from batch-, token-, or channel-level statistics. A recent study \citep{zhu2025transformers} 
introduces the Dynamic Tanh (DyT) function (\eqrefc{eq:DyT}), where $\alpha$ is a learnable parameter. This design is motivated by the observation that Layer Normalization often produces an S-shaped input-output mapping in practice. The saturating nature of the tanh function squashes extreme activations, thereby fulfilling a role analogous to the re-centering and re-scaling effects of normalization layers. 
\vspace{-0.04em}
\begin{equation}
    \mathrm{DyT}(x) = \gamma * \tanh(\alpha x) + \beta
    \label{eq:DyT}
\end{equation}
\vspace{-0.04em}
While DyT has shown similar empirical performance to normalization layers across various Transformer architectures, a comprehensive analysis of the design space for these statistics-free operators remains missing. 
In this work, we target at the optimal form of the point-wise function as normalization replacement.
We identify the function properties crucial for convergence and performance, and then we introduce \methodname{}, a point-wise function consistently surpassing normalization layers rather than merely matching their performance.

\begin{figure}[t]
    \centering
    \includegraphics[width=\textwidth]{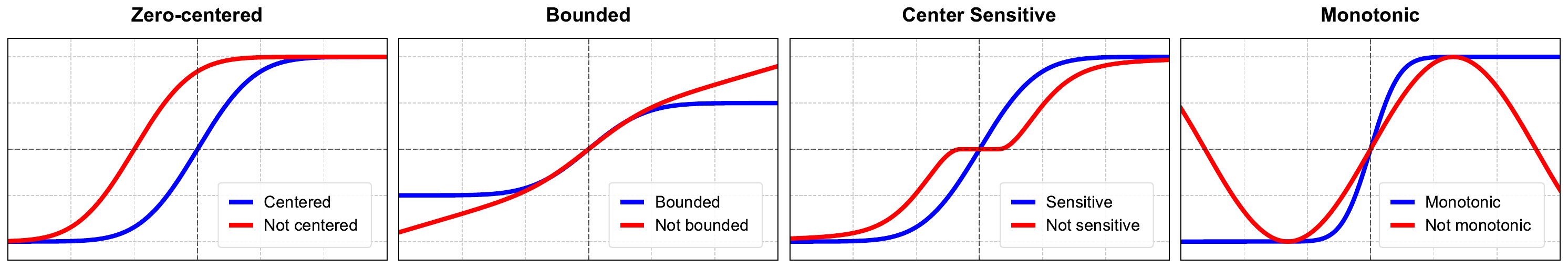}
    \caption{\textbf{Key properties of point-wise function.}
    The four properties: \textit{zero-centeredness}, \textit{boundedness}, \textit{center sensitivity}, and \textit{monotonicity} collectively characterize functional behavior on activations and influence training dynamics. Blue curves represent functions that satisfy each property, while red curves violate them.}
    \label{fig:activation_properties}
\end{figure}

\section{Function Property Analysis}
Training Transformers without normalization requires understanding the factors that make a point-wise function stable and effective as a replacement. In this section, we examine four essential properties: \textit{zero-centeredness}, \textit{boundedness}, \textit{center sensitivity}, and \textit{monotonicity} (see \figref{fig:activation_properties}). These properties collectively characterize the fundamental shape of point-wise functions and their behavior on activations. By isolating the impact of each property, we explore its influence on optimization and final performance.

To investigate these properties, we replace each normalization layer with a point-wise function of the form:
\vspace{-.05cm}
\begin{equation}
    y = \gamma \cdot f(\alpha x) + \beta,
\vspace{-.2cm}
\end{equation}

where $f(\cdot)$ denotes the chosen base function with learnable $\alpha$ rescaling the input. $\gamma$ and $\beta$ are affine parameters, similar to those in normalization layers. We begin with three base functions: $\tanh(x)$, $\mathrm{erf}(x)$, and $\arctan(x)$. In subsequent experiments, we modify these functions with controlled transformations to examine the impact of each property. All experiments are conducted with ViT-Base \citep{dosovitskiy2020image}, and top-1 accuracy on ImageNet-1K \citep{deng2009imagenet} is reported. In \appendixref{appendix:property_analysis_details}, we provide more detailed training results.


\subsection{Zero-centeredness}
\vspace{-.1cm}
\textit{Zero-centeredness} means that the function’s outputs are balanced around zero, with positive and negative values of similar magnitude and symmetry. Because normalization layers inherently recenter activations to the origin for stabilizing gradients, maintaining this property could reduce internal covariate shifts and promote smoother gradient flow during training.

\paragraph{Setup.}
Under the ViT setup, we manipulate the centering of the functions. For each base function, we consider two types of shifts: horizontal and vertical, defined in \eqrefc{eq:shift}. In this form, $\lambda_{\text{horiz}}$ and $\lambda_{\text{vert}}$ respectively denote the magnitudes of horizontal and vertical shifts. For both types of shifts, we vary $\lambda$ over $\{\pm \tfrac{1}{2}, \pm 1, \pm 2\}$ to examine how increasing deviation from zero-centeredness affects the function’s behavior. All other training settings remain unchanged.
\begin{equation}
\vspace{-.1cm}
    f_{\text{horiz}}(x) = f(x + \lambda_{\text{horiz}}), \quad
    f_{\text{vert}}(x) = f(x) + \lambda_{\text{vert}},
    \label{eq:shift}
\vspace{-0.5em}
\end{equation}

\paragraph{Results.}
As shown in \tabref{tab:zero_center}, the results are consistent across different base functions: for horizontal shifts, performance remains largely comparable to the zero-centered base function when $|\lambda_{\text{horiz}}| \leq 0.5$. However, as $|\lambda_{\text{horiz}}|$ increases, performance gradually degrades, and training diverges when $|\lambda_{\text{horiz}}| \geq 2$. Similarly, vertical shifts consistently lead to a decline in performance as $|\lambda_{\text{vert}}|$ grows with training failure once $|\lambda_{\text{vert}}| \geq 2$. These results show that \textit{zero-centeredness} is a requirement for stable convergence and effective training.

\begin{table}[h]
\vspace{0.5em}
\centering
\tablestyle{8pt}{1.25}
\begin{tabular}{lcccccccccc}
\toprule
function & shift type & -2 & -1 & -0.5 & -0.1 & $\lambda=0$ & +0.1 & +0.5 & +1 & +2 \\
\midrule
\multirow{2}{*}{$\mathrm{erf}(x)$} 
    & horizontal & $\times$ & 82.0\% & 82.5\% & 82.6\% & 82.6\% & \textbf{82.7\%} & 82.5\% & 82.1\% & $\times$ \\
    & vertical   & $\times$ & 81.8\% & 82.3\% & 82.4\% & \textbf{82.6\%} & 82.5\% & 82.3\% & 81.6\% & $\times$ \\
\midrule
\multirow{2}{*}{$\tanh(x)$ }
    & horizontal & $\times$ & 82.1\% & 82.5\% & \textbf{82.6\%} & 82.5\% & \textbf{82.6\%} & 82.4\% & 82.2\% & $\times$ \\
    & vertical   & $\times$ & 81.5\% & 81.9\% & 82.4\% & \textbf{82.5\%} & 82.3\% & 81.9\% & 81.4\% & $\times$ \\
\midrule
\multirow{2}{*}{$\arctan(x)$}
    & horizontal & $\times$ & 81.9\% & 82.3\% & 82.3\% & 82.3\% & \textbf{82.4\%} & 82.2\% & 82.0\% & $\times$ \\
    & vertical   & $\times$ & 81.4\% & 81.9\% & 82.2\% & \textbf{82.3\%} & \textbf{82.3\%} & 82.0\% & 81.2\% & $\times$ \\
\midrule
\end{tabular}
\caption{\textbf{Results of zero-centeredness on ViT-Base.} Horizontal shift corresponds to modifying the input as $f(\alpha x \pm \lambda)$, while vertical shift adds or subtracts a constant to the output as $f(\alpha x) \pm \lambda$. “$\times$” indicates training failure.}
\label{tab:zero_center}
\vspace{-0.5em}
\end{table}

\vspace{-.3cm}
\subsection{Boundedness} 
\label{sec:boundedness}

\textit{Boundedness} refers to the property of a function whose output is constrained within a finite range. Formally, a function \( f(\cdot) \) is bounded if there exist constants \( a, b \in \mathbb{R} \) such that \( a \le f(x) \le b \) for all \( x \) in its domain. This ensures that activations remain finite and do not accumulate variance across layers. Unbounded functions, in contrast, may induce signal explosion and gradient instability.

\paragraph{Setup.}
Under the same ViT setup, we study the role of \textit{boundedness} with two methods.
Firstly, we select three inherently unbounded S-shaped functions (e.g., $\mathrm{arcsinh}(x)$) and compare them with their clamped versions shown in \eqrefc{eq:Boundedness_Clamp}, 
where $f_{u}(x)$ denotes the unbounded point-wise function, and $\lambda$ is a chosen value specifying the clipping range. 
\begin{equation}
\vspace{1em}
y = \mathrm{clip}(f_{u}(x), -\lambda_{u}, \lambda_{u}),
\label{eq:Boundedness_Clamp}
\vspace{0.2em}
\end{equation}
Secondly, we gradually transition bounded functions (e.g., $\mathrm{erf}(x)$) toward unbounded linear form, defined in \eqrefc{eq:Boundedness_Mix}, 
where $f_{b}$ denotes a bounded point-wise function, and $\lambda$ controls how quickly the function becomes unbounded. We vary $\lambda_{u}$ over $\{0.5, 0.8, 1.0, 2.0, 3.0, 5.0\}$ in the first method and $\lambda_{b}$ over $\{0.01, 0.1, 0.5\}$ for the second. The original unmodified function is also included as a baseline.
\begin{equation}
y = (1-\lambda)f_{b}(x) + \lambda_{b} x, \quad \lambda_{b} \in (0,1).
\vspace{-0.8em}
\label{eq:Boundedness_Mix}
\end{equation}

\begin{figure}[!t]
\centering

\begin{minipage}[t]{0.62\textwidth}
\vspace{0pt}
\centering
\includegraphics[width=0.82\textwidth]{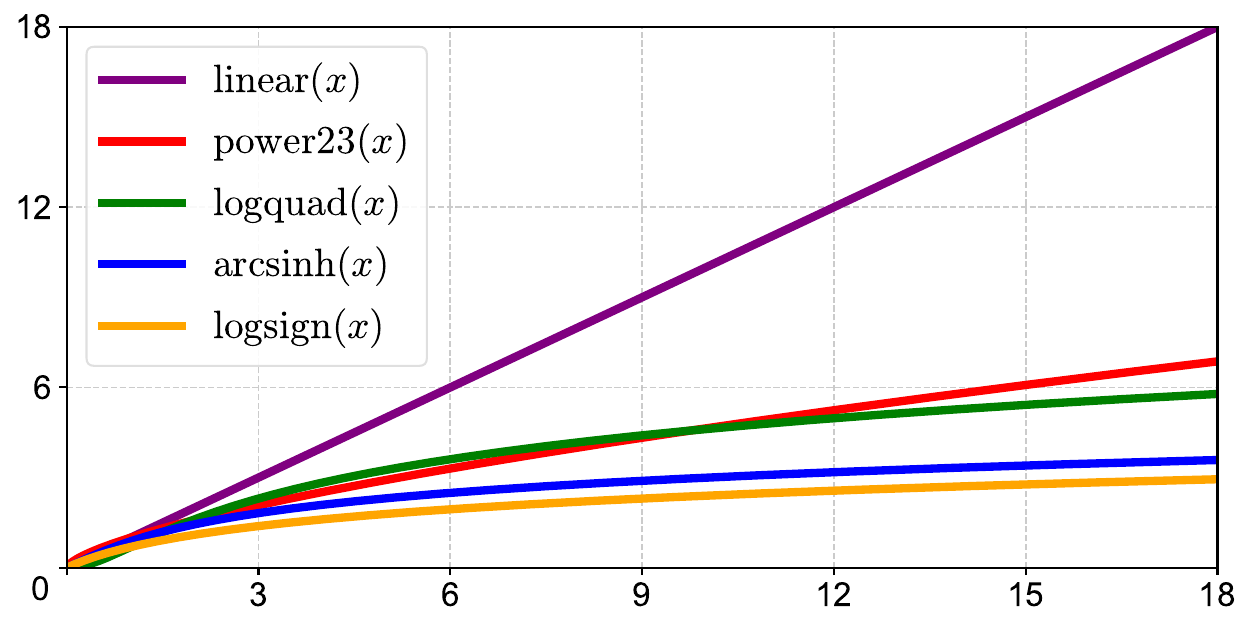}
\caption{\textbf{Visualization of several unbounded point-wise functions on the positive half-axis, illustrating their different growth rates. }$arcsinh(x)$ refers to its standard analytical form. The remaining functions are defined as 
$\mathrm{linear}(x)=x$,
$\mathrm{power23}(x)=\operatorname{sign}(x) \cdot x^{\frac{2}{3}}$,
$\mathrm{logsign}(x)=\operatorname{sign}(x)\ln(|x|+1)$, 
$\mathrm{smoothsign}(x) = \frac{x}{1 + |x|}$, 
and $\mathrm{logquad}(x)=\operatorname{sign}(x)\ln(x^{2}+1)$. Among them, $\mathrm{logquad}(x)$ shows the fastest growth that still ensures stable convergence.}
\label{fig:unbounded_functions}
\end{minipage}
\hfill
\begin{minipage}[t]{0.35\textwidth}
\vspace{0.58em}
\centering
\tablestyle{4pt}{1.30}
\begin{tabular}{lccc}
\toprule
$\lambda_u$ & $\mathrm{arcsinh}(x)$ & $\mathrm{logsign}(x)$ & $\mathrm{linear}(x)$\\
\midrule
$-$ & 82.2\% & 82.2\% & $\times$ \\
0.5 & 82.3\% & \textbf{82.4\%}  & 82.1\% \\
0.8 & 82.3\% & \textbf{82.4\%} & \textbf{82.2\%} \\
1.0 & \textbf{82.4\%} & \textbf{82.4\%} & \textbf{82.2\%} \\
2.0 & \textbf{82.4\%} & \textbf{82.4\%} & 82.1\%\\
3.0 & \textbf{82.4\%} & 82.3\%  & 82.1\% \\
5.0 & 82.3\% & 82.3\%  & 82.0\% \\
\midrule
\end{tabular}
\vspace{0.30em}
\captionof{table}{\textbf{Results of clamping for boundedness on ViT-Base.} Clipped version of unbounded functions consistently achieves better performance than unbounded baselines. “$-$” denotes the original unmodified function. “$\times$” indicates training failure. }
\label{tab:boundedness_results}
\end{minipage}
\end{figure}

\paragraph{Results.}
For the first method, among the three unbounded functions in \tabref{tab:boundedness_results}, only $\mathrm{arcsinh}(x)$ and $\mathrm{logsign}(x)$ converge effectively, while $\mathrm{linear}(x)$ does not. For the convergent functions, their clipped versions consistently outperform the unbounded baselines across all tested $\lambda$ values. These results indicate that incorporating \textit{boundedness} can improve optimization and result in better performance. For the second, as shown in \tabref{tab:extra_boundedness_results}, the results are consistent with clipping the intrinsic unbounded functions: the unbounded variant yields slightly lower accuracy than the bounded baseline. 

\begin{table}[h]
\vspace{0.5em}
\centering
\tablestyle{10pt}{1.15}
\centering
\begin{tabular}{lcccc}
\toprule
$\lambda_b$ & $\mathrm{erf}(x)$ & $\tanh(x)$ & $\mathrm{arctan}(x)$ & $\mathrm{isru}(x)$ \\
\midrule
$-$ & \textbf{82.6\%} & \textbf{82.5\%} & \textbf{82.3\%}  & \textbf{82.2\%}\\
0.01 & 82.4\% & 82.4\% & 82.1\% & \textbf{82.2\%}\\
0.1 & 82.3\% & 82.3\% & 82.1\% & 82.1\% \\
0.5 & $\times$ & $\times$ & $\times$ & $\times$\\
\midrule
\end{tabular}
\vspace{-0.1em}
\caption{\textbf{Results of removing boundedness on ViT-Base.} Performance decreases as the function is less bounded. “$-$” denotes the original function without modification and “$\times$” donotes training failure.
}
\label{tab:extra_boundedness_results}
\end{table}

\paragraph{Limitation of growth rate.} From \tabref{tab:boundedness_results} and \tabref{tab:extra_boundedness_results}, we observe that there is an upper limit on their acceptable growth rate. Large growth rates often lead to training failure. To determine this limit, we evaluate a family of inherently unbounded functions with varying growth rates, as illustrated in \figref{fig:unbounded_functions}.
Among them, $\mathrm{logquad}(x)$ exhibits the fastest growth that still allows training convergence (see \tabref{tab:limit_growrate}). Functions with faster growth, such as $\mathrm{linear}(x)$ and $\mathrm{power23}(x)$, tend to cause optimization divergence in the early stages of training. This failure occurs because rapidly growing functions fail to suppress variance effectively, leading to large gradient norms at the start of optimization.

\begin{table}[h!]
\vspace{0.5em}
\centering
\tablestyle{4.8pt}{1.25}
\begin{tabular}{ccccc}
\toprule
$\mathrm{logsign}(x)$ & $\mathrm{arcsinh}(x)$ &  $\mathrm{logquad}(x)$ & $\mathrm{power23}(x)$ & $\mathrm{linear}(x)$ \\
\midrule
82.2\% & 82.2\% & 82.1\% & $\times$ & $\times$ \\
\midrule
\end{tabular}
\caption{\textbf{Results of unbounded functions with different growth rates on ViT-Base.} Point-wise functions have a growth rate upper bound, with $\mathrm{logquad}(x)$ being the fastest function that still converges. “$\times$” indicates training failure. }
\label{tab:limit_growrate}
\end{table}

\vspace{-.3cm}

\subsection{Center Sensitivity} 
We use \textit{center sensitivity} to characterize how quickly a point-wise function becomes responsive to input variations around zero. Without \textit{center sensitivity}, a function is locally flat around the origin, returning zero or near-zero over a finite interval. The region around zero is particularly important, as most activations tend to concentrate near the origin during training. Consequently, the responsiveness of a function in this area directly influences how effectively small signals can propagate through the network.

\paragraph{Setup.}
Since \textit{center sensitivity} is difficult to isolate independently, we approximate it using a controllable near-zero inactive region.  
Under the same ViT setup, we modify each base function to incorporate a symmetric flat region around the origin with a sensitivity scale $\lambda>0$ to control the extent of this region.
Specifically, for inputs in the range $x \in [-\lambda, \lambda]$, we enforce $f(x)=0$ and smoothly shift the positive and negative parts outward for $|x| > \lambda$ to ensure continuity at the boundaries. A smaller $\lambda$ results in a narrower flat region and higher sensitivity near zero, while a larger $\lambda$ leads to lower sensitivity. We vary $\lambda$ over $\{0.1, 0.5, 1.0, 2.0, 3.0\}$ across three base functions.

\paragraph{Results.}
As shown in \tabref{tab:flat_zone_width}, the best performance is achieved at $\lambda=0$. As $\lambda$ increases, the performance consistently degrades. This trend is not very clear when $\lambda \le 0.5$, but once $\lambda$ exceeds 1.0, the degradation becomes much more obvious. Finally, when $\lambda \geq 3.0$, the training process diverges at an early stage. 

\begin{table}[h!]
\vspace{0.5em}
\centering
\tablestyle{6pt}{1.25}
\begin{tabular}{lcccccc}
\toprule
function & $\lambda=0$ & 0.1 & 0.5 & 1.0 & 2.0 & 3.0 \\
\midrule
$\mathrm{erf}(x)$  & \textbf{82.6\%} & 82.5\% & 82.5\% & 82.1\% &  81.3\% & $\times$  \\
$\tanh(x)$  & \textbf{82.5\%} & \textbf{82.5\%} & 82.4\% & 82.1\% & 81.1\% & $\times$  \\
$\arctan(x)$ & \textbf{82.3\%} & \textbf{82.3\%} & 82.1\% & 81.8\% & 80.9\% & $\times$  \\
\midrule
\end{tabular}
\caption{\textbf{Results of center sensitivity ($\lambda$) on ViT-Base.} “$\times$” indicates training failure. The best performance is achieved when no flat region is given, showing the importance of center sensitivity.}
\label{tab:flat_zone_width}
\end{table}


\subsection{Monotonicity}
\textit{Monotonicity} ensures a function’s output consistently increases (or decreases) as the input increases, preserving the relative order of inputs throughout the transformation. Non-monotonic functions may disrupt the relative ordering of activations. Furthermore, since a non-monotonic function necessarily has regions where its derivative changes sign, it may also produce flipped gradient signals during training. 

\paragraph{Setup.}
Each base function selected can serve as the monotonically increasing case, while its negated counterpart is defined as $f_{\text{neg}}(x) = -f(x)$, representing the monotonically decreasing variant. As non-monotonic comparisons, we include
 hump-shaped functions and oscillatory functions (see \figref{fig:monotonicity_functions}) to examine how violations of \textit{monotonicity} influence the training performance. 
To control potential confounding factors, we rescale each function so that its output range matches that of the monotonic functions. After rescaling, all functions are aligned in terms of \textit{zero-centeredness}, \textit{boundedness}, and \textit{center sensitivity}.

\begin{figure}[h]
    \centering
    \begin{minipage}[b]{0.42\textwidth}
        \centering
        \includegraphics[width=1\textwidth]{figures/monotonicity.pdf}
        \caption{\textbf{Visualization of point-wise functions with different monotonicity behaviors.} $\mathrm{erf}(x)$ and $\sin(x)$ refer to their standard form. The remaining functions are defined as 
        $\mathrm{negerf}(x)=-\mathrm{erf}(x)$,
        $\mathrm{dampx}(x)=\tfrac{2x}{1+x^2}$,
        $\mathrm{dampexp}(x)=2.72 x \cdot e^{-|x|}$}
        \label{fig:monotonicity_functions}
    \end{minipage}
    \hfill
    \begin{minipage}[b]{0.54\textwidth}
        \centering
        \tablestyle{7pt}{1.45} 
        \raisebox{25mm}{%
            \begin{subtable}[t]{0.56\textwidth}
            \centering
            \captionsetup{justification=centering} 
            \begin{tabular}{lcc}
            \toprule
            function & $f(x)$ & $f_{\text{neg}}(x)$ \\
            \midrule
            $\mathrm{erf}(x)$  &  82.6\%   &   82.5\%    \\
            $\tanh(x)$     &   82.5\%   &  82.5\%     \\
            $\arctan(x)$   &   82.3\%   &  82.2\%     \\
            \midrule
            \end{tabular}
            \caption{Monotonic}
        \end{subtable}
        \begin{subtable}[t]{0.40\textwidth}
            \centering
            \captionsetup{justification=centering} 
            \begin{tabular}{lc}
            \toprule
            function & $f(x)$ \\
            \midrule
            $\mathrm{sin}(x)$      &    81.6\%   \\
            $\mathrm{dampx}(x)$    &    80.7\%   \\
            $\mathrm{dampexp}(x)$  &    81.2\%    \\
            \midrule
            \end{tabular}
            \caption{Non-monotonic}
        \end{subtable}}

        \captionof{table}{\textbf{Results of monotonicity on ViT-Base.} Monotonic functions consistently achieve better performance than their negated versions and other non-monotonic functions, whether hump-shaped or oscillatory. This identifies monotonicity as a key property for effective learning.
        }
        \label{tab:monotonicity_results}
    \end{minipage}
    \vspace{-0.5em}
\end{figure}

\paragraph{Results.}
As shown in \tabref{tab:monotonicity_results}, both increasing and decreasing monotonic functions train stably and achieve high accuracy. In contrast, non-monotonic functions, whether hump-shaped or oscillatory, consistently perform worse than monotonic functions and lead to a clear drop in final accuracy. These results highlight \textit{monotonicity} as a key property for point-wise functions to ensure effective learning.



\begin{table}[t]
\centering
\tablestyle{15pt}{1.45}
\begin{tabular}{l l ccc}
\toprule
\multirow{2}{*}{function} &
\multirow{2}{*}{alias} &
\multicolumn{1}{c}{top-1 acc $\uparrow$} &
\multicolumn{2}{c}{FID $\downarrow$} \\
\cmidrule(lr){3-3}\cmidrule(lr){4-5}
& & ViT-Base & DiT-B/4 & DiT-L/4 \\
\midrule
-- & LayerNorm & 82.3\% & 64.93 & 45.91 \\
\midrule
$2\pi^{-1/2}\!\int_{0}^{x} e^{-t^{2}}\,dt$ 
    & $\mathrm{erf}(x)$ 
    & \textbf{82.8\%} & \textbf{63.23} & \textbf{43.94} \\

$(e^{x}-e^{-x})(e^{x}+e^{-x})^{-1}$ 
    & $\tanh(x)$ 
    & 82.6\% & 63.71 & 45.48 \\

$\sin (\mathrm{clip}(x,\,-\tfrac{\pi}{2},\,\tfrac{\pi}{2}))$
    & $\mathrm{satursin}(x)$ 
    & 82.6\% & 63.90 & 44.83 \\

$\mathrm{clip}~\!(\ln(x + \sqrt{x^{2} + 1}),\, -1,\, 1)$
    & $\mathrm{arcsinh_{clip}}(x)$ 
    & 82.5\% & 64.72 & 45.48 \\

$x(x^{2}+1)^{-1/2}$
    & $\mathrm{isru}(x)$ 
    & 82.3\% & 65.72 & 45.93 \\

$\operatorname{sign}(x)\,((1-e^{-\sqrt{|x|}}))$
    & $\mathrm{exproot}(x)$
    & 82.4\% & 65.20 & 46.91 \\

$\mathrm{clip}(x,\,-1,\,1)$
    & $\operatorname*{linear_{\text{clip}}}(x)$
    & 82.3\% & 66.08 & 45.49 \\

$-\operatorname{sign}(x)\,(e^{-|x|}-1)$
    & $\mathrm{expsign}(x)$
    & 82.2\% & 64.85 & 45.82 \\

$\mathrm{clip}~\!(\operatorname{sign}(x)\,\ln(|x|+1),\, -1,\, 1)$
    & $\operatorname*{logsign_{\text{clip}}}(x)$
    & 82.4\% & 65.59 & 46.34 \\

$x(\sqrt{x^{2}+1}+1)^{-1}$
    & $\mathrm{relsign}(x)$
    & 82.3\% & 68.42 & 48.33 \\

$\arctan(x)$
    & $\arctan(x)$
    & 82.4\% & 67.07 & 46.62 \\

$x(1+|x|)^{-1}$
    & $\mathrm{smoothsign}(x)$
    & 82.4\% & 68.84 & 47.29 \\

$\mathrm{clip}~\!(\operatorname{sign}(x)\,\ln(x^{2}+1),\, -1,\, 1)$
    & $\operatorname*{logquad_{\text{clip}}}(x)$
    & 82.2\% & 65.92 & 47.12 \\

$\mathrm{clip}~\!(\operatorname{sign}(x)\,x^{2/3},\, -1,\, 1)$
    & $\operatorname*{power23_{\text{clip}}}(x)$
    & 82.1\% & 66.11 & 46.47 \\

$\operatorname{sign}(x)\,\ln(|x|+1)\, (\ln(|x|+1)+1)^{-1}$
    & $\mathrm{saturlog}(x)$
    & 81.8\% & 68.23 & 47.44 \\

$x^{3}(|x|^{3}+1)^{-1}$
    & $\mathrm{cubsign}(x)$
    & 81.4\% & 70.22 & 49.16 \\
\midrule

\end{tabular}

\caption{\textbf{Top-1 accuracy on ViT-Base and image generation quality (FID) on DiT-B/4 and DiT-L/4.} Different functions show noticeable differences in performance. Among all the point-wise functions and LayerNorm, $\mathrm{erf}(x)$ shows the best performance in both top-1 accuracy and FID. Visualization of each function is included in \appendixref{appendix:function_search_details}.} 
\label{tab:emprical_search_results}

\vspace{-1.3em}

\end{table}

\section{Function Search}
From the previous section, we observe that functions that are near \textit{zero-centered}, \textit{bounded}, \textit{center-sensitive} (responsive to input variations around zero), and \textit{monotonic} (increasing or decreasing) tend to yield better optimization performance. Building upon these insights, we start to construct our function set from widely used scalar functions and cumulative distribution functions (CDFs), including polynomial, rational, exponential, logarithmic, and trigonometric forms. We then generate variants via simple transformations such as translation, scaling, mirroring, rotation, and clipping. Functions that satisfy our four function properties after these transformations are retained as the candidate subset used in the search. 
For example, we transform the unbounded function $\mathrm{arcsinh}(x)$ by clipping it to the range $[-1, 1]$, limiting it to a finite range and conforming to all four principles.
In \appendixref{appendix:function_search_details}, we provide further details about how we obtain these candidate functions. Within this set, we evaluate their performance, and \methodname{} emerges as the most effective function.

\paragraph{Setup.} We conduct an empirical search on two representative vision architectures: Vision Transformer (ViT-Base) \citep{dosovitskiy2020image} and Diffusion Transformer (DiT-B/4 and DiT-L/4) \citep{peebles2023scalable}. Models are trained on ImageNet-1K \citep{deng2009imagenet} under their default training settings. For ViT, model performance is measured using top-1 accuracy on the ImageNet-1K validation set. For DiT, we follow the standard ImageNet reference batch evaluation and report the Fréchet Inception Distance (FID) as the metric.

\begin{figure}[t]
    \centering
    \includegraphics[width=\textwidth]{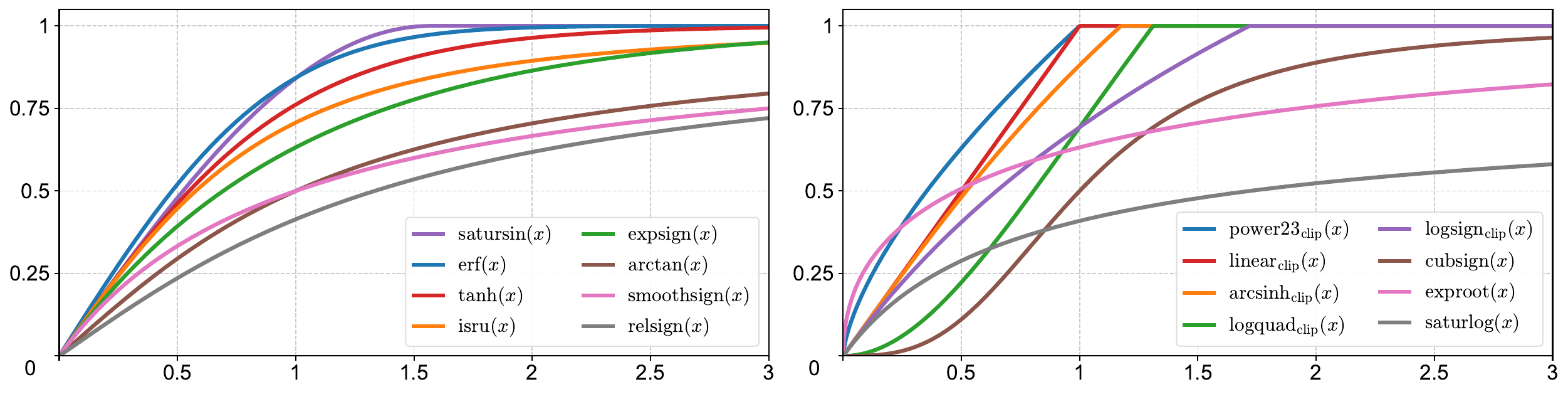}
    \caption{
\textbf{Visualization of candidate point-wise functions on the positive half-axis.}
All functions are self-symmetric with respect to the origin.
} \label{fig:bounded_functions}
\end{figure}

\paragraph{Formulation.} We quantitatively evaluate a set of functions under the constraint of our function properties, as illustrated in \figref{fig:bounded_functions}. Each point-wise function is instantiated in a unified form in \eqrefc{eq:pointwise}, where $f(\cdot)$ denotes a candidate point-wise function, with learnable parameter $s$ and $\alpha$ recentering and rescaling the input. The parameters $\gamma$ and $\beta$ follow the same role as in standard normalization layers. We introduce a learnable shift parameter $s$, as it improves the final performance to varying degrees across different functions. Detailed ablation results on the effect of $s$ are provided in \secref{subsec:ablation_s}. 

\vspace{-.3cm}
\begin{equation}
y = \gamma * f(\alpha x + s) + \beta,
\label{eq:pointwise}
\end{equation}

\newpage

\paragraph{Quantitative evaluation.} As shown in \tabref{tab:emprical_search_results}, even though these S-shaped functions appear highly similar in form, their empirical training results show noticeable differences in final performance. Among all the point-wise functions, $\mathrm{erf}(x)$ with the introduced transformations stands out as the best-performing function, consistently surpassing all other candidates and the baseline normalization layers.


\section{Dynamic erf (\methodname{})}
From the search, we identify $\mathrm{erf}(x)$ 
as the most performant point-wise function.
The error function $\mathrm{erf}(\cdot)$ is closely related to the cumulative distribution function (CDF) of a standard Gaussian distribution. Specifically, $\mathrm{erf}(x)$ can be defined by \eqrefc{eq:erf}. In our setup, $\mathrm{erf}(x)$ is in the form augmented with learnable parameters, which we introduce as \methodname{}, \textbf{D}ynamic \textbf{erf}. Given an input tensor $x$, a \methodname{} layer is defined in \eqrefc{eq:derf}, where both the shift $s$ and the scale $\alpha$ are learnable scalars. $\gamma$ and $\beta$ are learnable per-channel vectors. To integrate \methodname{} into a transformer-based architecture, we replace each normalization layer with a corresponding \methodname{} layer. In particular, the pre-attention, the pre-FFN, and the final normalization layers are all substituted in a one-to-one manner, ensuring consistent incorporation of \methodname{} across the entire model.

\vspace{-.3cm}
\begin{equation}
\mathrm{erf}(x) = \frac{2}{\sqrt{\pi}} \int_{0}^{x} e^{-t^2} dt
\label{eq:erf}
\end{equation}
\vspace{-.1cm}
\begin{equation}
\mathrm{Derf}(x) = \gamma\ * \mathrm{erf}(\alpha x + s) + \beta
\label{eq:derf}
\end{equation}
\vspace{-.7cm}

\vspace{0.1em}


\paragraph{Parameter initialization.}
We initialize $\gamma$ to an all-one vector and $\beta$ to an all-zero vector following the same strategy as in standard normalization layers. For the additional scalar parameters introduced by \methodname{}, the scaling parameter $\alpha$ is initialized to $0.5$, while the shift parameter $s$ is initialized to $0$.  Unless otherwise specified, these initialization settings are adopted throughout all experiments.

\section{Experiments}
We evaluate the effectiveness of \methodname{} across various transformer-based and a few other modern architectures. For each model, we replace the original normalization layers with DyT and \methodname{}, following the standard training and evaluation protocols, as detailed in \appendixref{appendix:experimental_settings}. Across all tested architectures, \methodname{} consistently achieves stronger performance over the baseline normalization methods and DyT. Besides each model’s default normalization, we also report results with other common normalization methods in \appendixref{appendix:additional_results}.

\paragraph{Vision Transformers.} We train ViT-Base and ViT-Large models \citep{dosovitskiy2020image} on ImageNet-1K \citep{deng2009imagenet} using LayerNorm (LN), DyT, and \methodname{} for comparison. \tabref{tab:vit_results} reports the top-1 classification accuracy. Compared to LN and DyT, \methodname{} achieves clearly higher top-1 accuracy.

\begin{table}[h!]
\vspace{0.3em}
\centering
\tablestyle{9pt}{1.25}
\begin{tabular}{lccccc}
\toprule
model & LN & DyT & \methodname{} & $\Delta_{\text{LN}}$ & $\Delta_{\text{DyT}}$ \\
\midrule
ViT-B & 82.3\% & 82.5\% & \textbf{82.8}\% & $\textcolor{ForestGreen}{\uparrow \text{0.5\%}}$ & $\textcolor{ForestGreen}{\uparrow \text{0.3\%}}$ \\
ViT-L & 83.1\% & 83.6\% & \textbf{83.8}\% & $\textcolor{ForestGreen}{\uparrow \text{0.7\%}}$ & $\textcolor{ForestGreen}{\uparrow \text{0.2\%}}$ \\
\midrule
\end{tabular}
\caption{\textbf{Supervised classification accuracy on ImageNet-1K.} \methodname{} achieves higher top-1 accuracy than both LN and DyT on different model sizes, demonstrating its effectiveness in vision transformer architectures.}
\label{tab:vit_results}
\vspace{-0.8em}
\end{table}

\newpage

\paragraph{Diffusion Transformers.}
We train three Diffusion Transformer (DiT) \citep{peebles2023scalable} models on ImageNet-1K \citep{deng2009imagenet}. Consistent with the original DiT setup, the affine parameters in the normalization layers are retained for class conditioning across LN, DyT, and \methodname{}. After training, we evaluate the FID scores using the standard ImageNet “reference batch” to measure image generation quality, as reported in \tabref{tab:dit_results}. \methodname{} achieves a clear improvement in FID compared to both LayerNorm and DyT.
\begin{table}[h!]
\vspace{0.3em}
\centering
\tablestyle{9pt}{1.25}
\begin{tabular}{lccccc}
\toprule
model & LN & DyT & \methodname{} & $\Delta_{\text{LN}}$ & $\Delta_{\text{DyT}}$ \\
\midrule
DiT-B/4 & 64.93 & 63.94 & \textbf{63.23} & $\textcolor{ForestGreen}{\downarrow \text{1.70}}$ & $\textcolor{ForestGreen}{\downarrow \text{0.71}}$ \\
DiT-L/4 & 45.91 & 45.66 & \textbf{43.94} & $\textcolor{ForestGreen}{\downarrow \text{1.97}}$ & $\textcolor{ForestGreen}{\downarrow \text{1.72}}$ \\
DiT-XL/2 & 19.94 & 20.83 & \textbf{18.92} & $\textcolor{ForestGreen}{\downarrow \text{1.02}}$ & $\textcolor{ForestGreen}{\downarrow \text{1.91}}$ \\
\midrule
\end{tabular}
\caption{\textbf{Image generation quality (FID) on ImageNet.} Lower FID indicates better image generation quality. \methodname{} achieves lower FID scores than both LN and DyT across all DiT model sizes.}
\label{tab:dit_results}
\vspace{-1.2em}
\end{table}

\paragraph{Speech models.} We train two wav2vec 2.0 Transformer models \citep{baevski2020wav2vec} on the LibriSpeech dataset \citep{panayotov2015librispeech} for speech representation learning. We report the final validation loss in \tabref{tab:wav2vec_results}. Compared to LayerNorm and DyT, \methodname{} yields lower validation loss on different model sizes.

\begin{table}[h!]
\vspace{0.3em}
\centering
\tablestyle{7.3pt}{1.25}
\begin{tabular}{lccccc}
\toprule
model & LN & DyT & \methodname{} & $\Delta_{\text{LN}}$ & $\Delta_{\text{DyT}}$ \\
\midrule
wav2vec 2.0 Base & 1.95 & 1.95 & \textbf{1.93} & $\textcolor{ForestGreen}{\downarrow \text{0.02}}$ & $\textcolor{ForestGreen}{\downarrow \text{0.02}}$ \\
wav2vec 2.0 Large & 1.92 & 1.91 & \textbf{1.90} & $\textcolor{ForestGreen}{\downarrow \text{0.02}}$ & $\textcolor{ForestGreen}{\downarrow \text{0.01}}$ \\
\midrule
\end{tabular}
\caption{\textbf{Speech pretraining validation loss on the LibriSpeech dataset.} \methodname{} achieves lower validation loss than both LN and DyT across two wav2vec~2.0 models, indicating its better representation quality.}

\label{tab:wav2vec_results}
\vspace{-0.8em}
\end{table}

\paragraph{DNA models.}
For the long-range DNA sequence modeling task, we pretrain the HyenaDNA model \citep{nguyen2023hyenadna} and the Caduceus model \citep{schiff2024caduceus} using the human reference genome from \citep{grch382013p13}. Model evaluation is conducted on the GenomicBenchmarks dataset \citep{grevsova2023genomic}. We report the averaged accuracy over all subtasks. As shown in \tabref{tab:dna_results}, \methodname{} surpasses both normalization layers and DyT in performance, demonstrating its robustness in genomic sequence modeling.

\begin{table}[h!]
\centering
\vspace{0.3em}
\tablestyle{7.7pt}{1.25}
\begin{tabular}{lccccc}
\toprule
model & Norm & DyT & \methodname{} & $\Delta_{\text{Norm}}$ & $\Delta_{\text{DyT}}$ \\
\midrule
Hyena & 85.2\% & 85.2\% & \textbf{85.7}\% & $\textcolor{ForestGreen}{\uparrow \text{0.5\%}}$ & $\textcolor{ForestGreen}{\uparrow \text{0.5\%}}$ \\
Caduceus & 86.9\% & 86.9\% & \textbf{87.3}\% & $\textcolor{ForestGreen}{\uparrow \text{0.4\%}}$ & $\textcolor{ForestGreen}{\uparrow \text{0.4\%}}$ \\
\midrule
\end{tabular}
\caption{\textbf{DNA classification accuracy on the GenomicBenchmarks dataset}, averaged over each subtask. Each model is evaluated with its default normalization layer (LN for Heyna, RMSNorm for Caduceus). \methodname{} consistently achieves higher accuracy than both normalization layers and DyT, indicating its effectiveness in DNA model.}
\label{tab:dna_results}
\end{table}

\newpage

\paragraph{Language models.}
We pretrain a GPT-2 (124M) model on the OpenWebText dataset and report the validation loss in \tabref{tab:llm_results}. For DyT and \methodname{}, we additionally finetune the initialization of the learnable parameter $\alpha$. We observe that \methodname{} achieves comparable performance to LN, while clearly outperforming DyT.

\begin{table}[h!]
\centering
\vspace{0.3em}
\tablestyle{11.3pt}{1.25}
\begin{tabular}{lccccc}
\toprule
model & LN & DyT & \methodname{} & $\Delta_{\text{LN}}$ & $\Delta_{\text{DyT}}$ \\
\midrule
GPT-2  & 2.94 & 2.97 & 2.94 & $\textcolor{gray}{\text{0.00}}$
 & $\textcolor{ForestGreen}{\downarrow \text{0.03}}$ \\
\midrule
\end{tabular}
\caption{\textbf{GPT-2 validation loss on the OpenWebText dataset.} \methodname{} matches the performance of LN while achieving lower validation loss than DyT. }
\label{tab:llm_results}
\end{table}


\subsection{Stronger Generalization or Better Fitting?}
Given \methodname{}'s superior performance, we aim to determine whether the gains arise from improved fitting capacity or stronger generalization. To this end, we compare the training loss of models respectively trained with normalization layers, DyT, and \methodname{}. 
Since lower training loss indicates stronger fitting ability,  
this comparison helps us assess whether \methodname{} improves optimization or enhances generalization.

\paragraph{Setup.} 
We compute training losses across diverse architectures and scales. To measure fitting capacity fairly, we do not use the loss during optimization, which is confounded by stochastic regularization (e.g., stochastic depth \citep{huang2016deep}) and train-time augmentations. Instead, after training, we switch to evaluation mode, disable stochastic depth (when present), adopt the test-time preprocessing pipeline, and compute the loss on the training set. This yields a fair estimate of each model's fitting capacity. In \appendixref{appendix:loss_calculation_details}, we provide the detailed procedure for computing the evaluation-mode training loss for each model.

\paragraph{Results.}
Across all architectures and scales, both \methodname{} and DyT result in higher training loss than normalization-based models, with \methodname{} generally yielding slightly lower training loss than DyT, as shown in \tabref{tab:train_loss_comparison}.
This consistent pattern indicates that neither \methodname{} nor DyT improves fitting capacity over normalization layers.

\begin{table}[h!]
\vspace{0.5em}
\tablestyle{15pt}{1.25}
\begin{tabular}{lccc}
\toprule
model & {Norm} & {\methodname{}} & DyT \\
\midrule
ViT-B      & \textbf{0.2623} & \underline{0.2681} & 0.2714 \\
ViT-L      & \textbf{0.2034} & \underline{0.2066} & 0.2083 \\
DiT-B      & \textbf{0.1531} & \underline{0.1533} & 0.1535 \\
DiT-L      & \textbf{0.1501} & \underline{0.1510} & 0.1518 \\
DiT-XL     & \textbf{0.1432} & \underline{0.1436} & 0.1440 \\
wav2vec 2.0 B & \textbf{1.8509} & \underline{1.8821} & 1.8946 \\
wav2vec 2.0 L & \textbf{1.8241} & \underline{1.8563} & 1.8641 \\
Hyena          & \textbf{1.1297} & \underline{1.1526} & 1.1631 \\
Caduceus        & \textbf{0.8917} & \underline{0.9129} & 0.9203 \\
GPT-2           & \textbf{2.9478} & \underline{2.9702} & 2.9822 \\
\midrule
\end{tabular}
\caption{\textbf{Evaluation-mode training loss of normalization layers (Norm), \methodname{}, and DyT after optimization.}    Bolded indicates the lowest loss, and underlined means the second-lowest loss.
Across all model architectures, the training loss follows the relation:  $\text{Norm}$~$<$~$\text{\methodname{}} < \text{DyT}$.
Both DyT and \methodname{} exhibit higher training loss than normalization layers, while \methodname{} achieves slightly lower loss than DyT.}
\label{tab:train_loss_comparison}
\end{table}

\paragraph{Discussion.}
Despite the reduced fitting capacity, \methodname{} delivers consistent performance gains across all evaluated tasks. We hypothesize that these gains arise primarily from both better generalization than normalization layers and stronger fitting capacity than DyT.

Firstly, point-wise functions promote stronger generalization. Although \methodname{} yields higher training loss, it achieves superior downstream performance, indicating that its benefits stem not from improved fitting but from enhanced generalization. This difference likely originates from the contrasting operational principles between normalization layers and point-wise functions. Normalization layers adapt their transformation based on training statistics, allowing them to dynamically fit activation distributions throughout training. In contrast, point-wise functions are controlled by only a small set of learnable scalar parameters (e.g., $\alpha$ for DyT and $\alpha$, $s$ for \methodname{}) that do not adapt to activation statistics after training. They apply the same transformation regardless of activation distribution. This limited adaptability constrains overfitting and effectively serves as an implicit regularizer, leading to improved generalization.

Secondly, \methodname{} exhibits stronger fitting power than DyT. It achieves lower training loss while retaining the implicit regularization of point-wise functions, combining higher fitting capacity with strong generalization to outperform both DyT and normalization-based models.

\section{Analysis}
In this section, we begin with two ablation studies examining the influence of the learnable shift parameter $s$ on the training results, followed by an analysis of an approximation of \methodname{}.

\subsection{Effect of s}  
\label{subsec:ablation_s}
\paragraph{Removing $\boldsymbol{s}$.} We investigate the effect of the learnable scalar parameter $s$ by removing it from the point-wise function. As shown in \tabref{tab:ablation_s}, introducing this learnable shift consistently improves the overall training performance, and the degree of improvement varies across different functions. The stronger results of $\mathrm{erf}(x)$ over $\tanh(x)$ indicate that \methodname{} surpasses DyT not only because of the shift $s$.

\begin{table}[h!]
\vspace{0.5em}
\centering
\tablestyle{8.6pt}{1.25}
\begin{tabular}{lcccc}
\toprule
\multicolumn{1}{l}{} &
\multicolumn{2}{c}{top-1 acc $\uparrow$} &
\multicolumn{2}{c}{FID $\downarrow$} \\
\cmidrule(lr){2-3} \cmidrule(lr){4-5}
function & without $s$ & with $s$ & without $s$ & with $s$ \\
\midrule
$\mathrm{erf}(x)$ & 82.6\% & \textbf{82.8\%} & 63.39 & \textbf{63.23} \\
$\mathrm{tanh}(x)$ & 82.5\% & \textbf{82.6\%} & 63.94 & \textbf{63.71} \\
$\mathrm{satursin}(x)$  & 82.4\% & \textbf{82.6\%} & 65.28 & \textbf{63.90} \\
$\mathrm{isru}(x)$       & 82.2\% & \textbf{82.3\%} & 66.14 & \textbf{65.72} \\
$\arctan(x)$     & 82.3\% & \textbf{82.4\%} & 67.41 & \textbf{67.07} \\
$\mathrm{arcsinh_{clip}}(x)$    & 82.4\% & \textbf{82.5\%} & 65.19 & \textbf{64.72} \\
\midrule
\end{tabular}
\caption{\textbf{Ablation study of $\boldsymbol{s}$.}
Top-1 accuracy on ViT-Base and FID score on DiT-B/4, comparing models with and without $s$. $s$ improves the overall training performance, while its effect varies across different point-wise functions.}
\label{tab:ablation_s}
\end{table}

\paragraph{Scalar vs. vector $\boldsymbol{s}$.}
We further examine whether using a per-channel vector parameter instead of a scalar $s$ leads to any performance improvement. As shown in \tabref{tab:scalar_vs_vector}, across all three point-wise functions, the choice between a scalar and a per-channel vector shows no significant impact on the final performance. Therefore, we adopt the scalar form of $s$ for efficiency and simplicity during training.

\begin{table}[h]
\vspace{0.5em}
\centering
\tablestyle{10pt}{1.25}

\begin{minipage}[t]{0.46\linewidth}
\centering
\begin{tabular}{lcc}
\toprule
function & vector & scalar \\
\midrule
$\mathrm{erf}(x)$     & \textbf{82.8\%} & \textbf{82.8\%} \\
$\mathrm{arctan}(x)$          & \textbf{82.5\%} & 82.4\% \\
$\mathrm{arcsinh_{clip}}(x)$ & \textbf{82.5\%} & \textbf{82.5\%} \\
\midrule
\end{tabular}
\caption{\textbf{Top-1 accuracy of scalar vs. vector $\boldsymbol{s}$ on ViT-Base.}
Using either a scalar or a per-channel vector for the parameter $s$ yields nearly identical performance.}
\label{tab:scalar_vs_vector}hu
\end{minipage}
\hfill
\begin{minipage}[t]{0.50\linewidth}
\centering
\begin{tabular}{lcccc}
\toprule
function & ViT-B & ViT-L & DiT-B & DiT-L \\
\midrule
$\tanh(x)$             & 82.6\% & 83.6\% & 63.71 & 45.48\\
$\tanh(\varepsilon x)$ & 82.7\% & 83.7\% & 63.88 & 45.13 \\
$\mathrm{erf}(x)$      & \textbf{82.8\%} & \textbf{83.8\%} & \textbf{63.23} & \textbf{43.94} \\
\midrule
\end{tabular}
\caption{\textbf{Top-1 accuracy of $\boldsymbol{\tanh(\varepsilon x)}$ on ViT and DiT.}
$\tanh(\varepsilon x)$ yields a comparable or slightly improved performance over $\tanh(x)$ but still remains below $\mathrm{erf}(x)$.}
\label{tab:approximation_erf}
\end{minipage}

\end{table}

\newpage

\subsection{Approximating \methodname{}}
Given the superior performance of $\mathrm{erf}(x)$ over $\tanh(x)$, we 
approximate $\mathrm{erf}(x)$ by scaling $\tanh(x)$ and examine whether this modification can lead to performance improvement. We introduce a fixed coefficient $\varepsilon$ and use $\tanh(\varepsilon x)$, where $\varepsilon$ is obtained by minimizing the following objective:
\begin{equation}
    \min_{\varepsilon} \int_{-\infty}^{+\infty} \big|\tanh(\varepsilon x) - \mathrm{erf}(x)\big|\,dx.
\end{equation}
The optimal value is found to be $\varepsilon \approx 1.205$. As shown in \tabref{tab:approximation_erf}, $\tanh(\varepsilon x)$ achieves a comparable or slightly improved performance over the original $\tanh(x)$, while still performing worse than $\mathrm{erf}(x)$. This indicates that simply scaling $\tanh(x)$ is insufficient to match the behavior or performance of $\mathrm{erf}(x)$.

\section{Related Work}

\paragraph{Normalization layers.}
Since the introduction of Batch Normalization (BN) \citep{ioffe2015batch}, various normalization methods have been proposed to better stabilize training. To address BN’s limitations with small batches, several alternatives \citep{salimans2016weight, wu2018group, yan2020towards, shen2020powernorm, singh2020filter} have been explored. In parallel, LayerNorm \citep{ba2016layer, nguyen2019transformers, xu2019understanding, xiong2020layer} and RMSNorm \citep{zhang2019root} were designed for RNN \citep{6795963} and Transformer architectures \citep{vaswani2017attention}. Task-specific variants \citep{ulyanov2016instance, wu2018group, shen2020powernorm} further adapt normalization to applications such as object detection and style transfer.

\paragraph{Mechanisms of normalization.}
A series of studies has investigated how normalization layers contribute to model convergence. From an optimization perspective, normalization stabilizes gradient flow \citep{balduzzi2017shattered, daneshmand2020batch, lubana2021beyond}, reduces sensitivity to initialization \citep{zhang2019fixup, de2020batch, shao2020normalization}, and implicitly tunes learning rates \citep{arora2018theoretical, tanaka2021noether}. It has also been shown to smooth the loss landscape \citep{santurkar2018does, bjorck2018understanding, karakida2019normalization} and reduce sharpness \citep{lyu2022understanding, dai2023crucial, mueller2023normalization}, promoting more stable optimization dynamics. Understanding these underlying functionalities provides valuable guidance for designing normalization-free training methods.

\paragraph{Normalization-free methods.}
Building on this understanding of normalization, recent work explores how to achieve stable convergence without normalization.
One line of work operates at the parameter and optimization level, using tailored initialization schemes \citep{zhang2019fixup, de2020batch, bachlechner2021rezero}, self-normalizing activations \citep{klambauer2017self}, weight normalization \citep{salimans2016weight, brock2021characterizing}, or adaptive gradient clipping \citep{brock2021high} to maintain stable gradient propagation. Another line of work modifies the architecture through structural simplifications \citep{he2023simplifying}, Softmax-only formulations \citep{jha2024aero}, and bounded convolutional operators \citep{liu2017deep, liu2018decoupled}. More recently, point-wise functions such as Dynamic Tanh \citep{zhu2025transformers} have been proposed, with theoretical analyses revealing their similarity to normalization operations \citep{stollenwerk2025mathematical}. Unlike previous methods that aim to match the performance of normalization layers, \methodname{} consistently delivers stronger performance across diverse models.

\section{Conclusion}
In this work, we demonstrate that well-designed point-wise functions do not merely match the performance of normalization layers, but can surpass them. By revisiting the design space of point-wise functions, we identify zero-centeredness, boundedness, center sensitivity, and monotonicity as four key properties that enable strong performance in Transformer-based models. Among the functions satisfying these properties, \methodname{} stands out as the most effective design: it consistently outperforms normalization-based methods and another notable point-wise function, DyT, across a wide range of modalities and tasks. Its simplicity and strong empirical performance make \methodname{} a compelling replacement for normalization layers in many Transformer architectures.

\clearpage

\subsection*{Acknowledgments}
We gratefully acknowledge the use of the Neuronic GPU computing cluster maintained by the Department of Computer Science at Princeton University. This work was substantially performed using Princeton Research Computing resources, a consortium led by the Princeton Institute for Computational Science and Engineering (PICSciE) and Research Computing at Princeton University. This work is also supported by the computational resources generously provided by Google’s TPU Research Cloud program. 

\bibliographystyle{plainnat}
\bibliography{main}

\appendix

\section*{\LARGE Appendix}

\vspace{.2cm}

\section{Property Analysis Details}
\label{appendix:property_analysis_details}

In this section, we provided detailed explanation and visualization on how different function properties affect model training.

\subsection{Zero-centeredness}
\begin{figure}[h!]
  \centering

  \begin{subfigure}{0.48\linewidth}
    \centering
    \captionsetup{justification=centering} 
    \includegraphics[width=\linewidth]{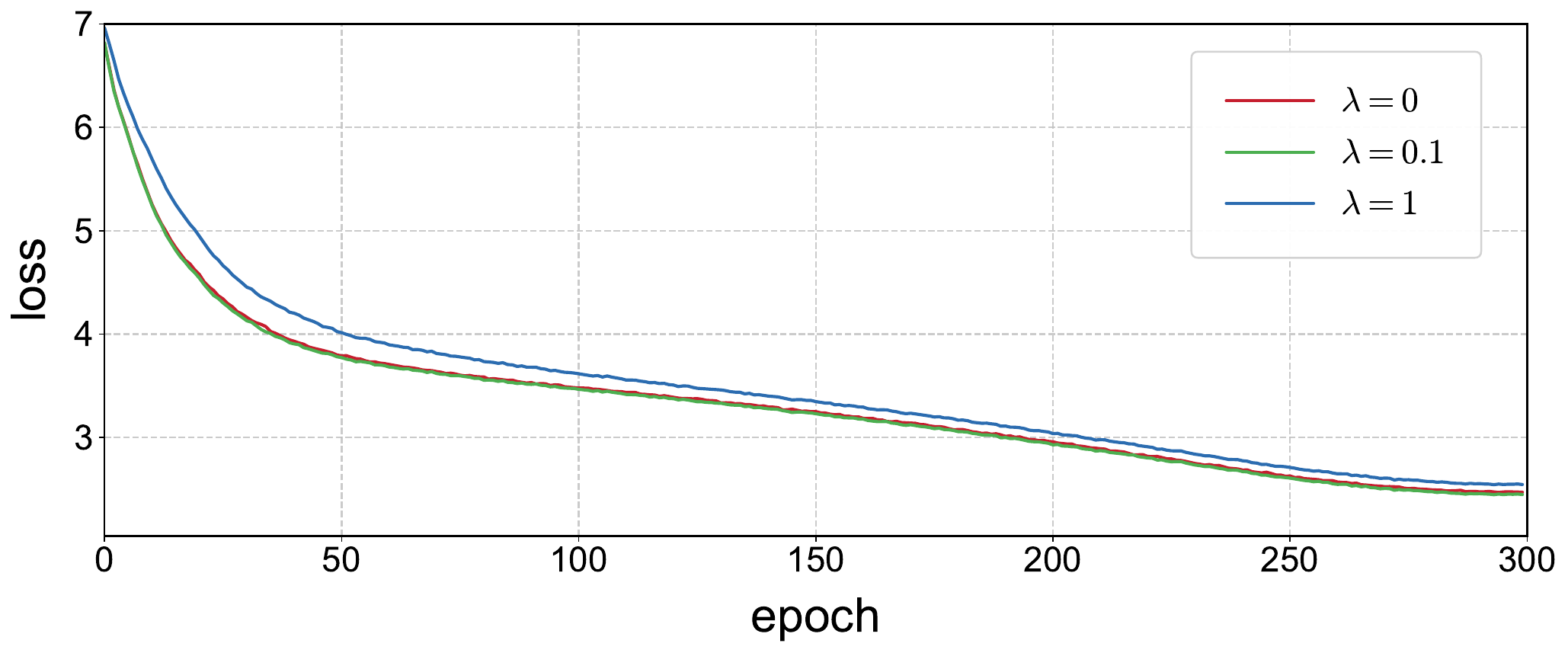}
    \caption{Horizontal shift}
  \end{subfigure}
  \hfill
  \begin{subfigure}{0.48\linewidth}
    \centering
    \captionsetup{justification=centering} 
    \includegraphics[width=\linewidth]{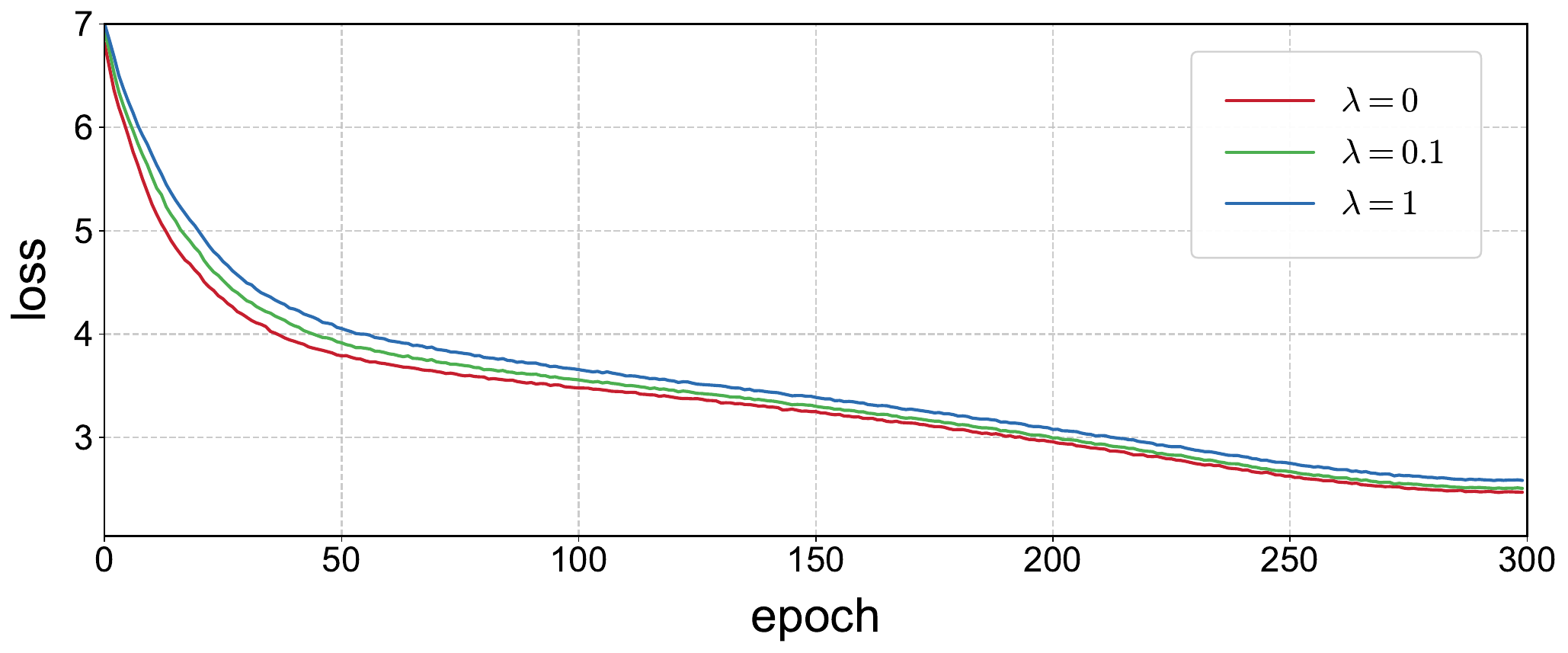}
    \caption{Vertical shift}
  \end{subfigure}

  \caption{\textbf{Training loss curve for horizontal and vertical shifts on the base point-wise function $\mathrm{erf}(x)$.} The trends are consistent with the patterns observed in top-1 accuracy on ImageNet-1K.}
  \label{fig:training_loss_zero_centeredness}
\end{figure}

We plot the training curves for \(\lambda_{\text{horiz}}\) and \(\lambda_{\text{vert}}\) with values \(\{0, 0.1, 1\}\) in \figref{fig:training_loss_zero_centeredness}. The trends are consistent with those observed in top-1 accuracy on ImageNet-1K. For horizontal shifts, the training loss with \(\lambda_{\text{horiz}} = 0.1\) nearly overlaps with that of \(\lambda_{\text{horiz}} = 0\), and even reaches a slightly lower loss. In contrast, vertical shifts exhibit a monotonic pattern: increasing \(\lambda_{\text{vert}}\) consistently raises the training loss, suggesting reduced fitting capacity under larger vertical shift.

\subsection{Center Sensitivity}
\begin{figure}[h!]
\centering

\begin{minipage}[t]{0.42\linewidth}
    \centering
    \includegraphics[width=0.85\linewidth]{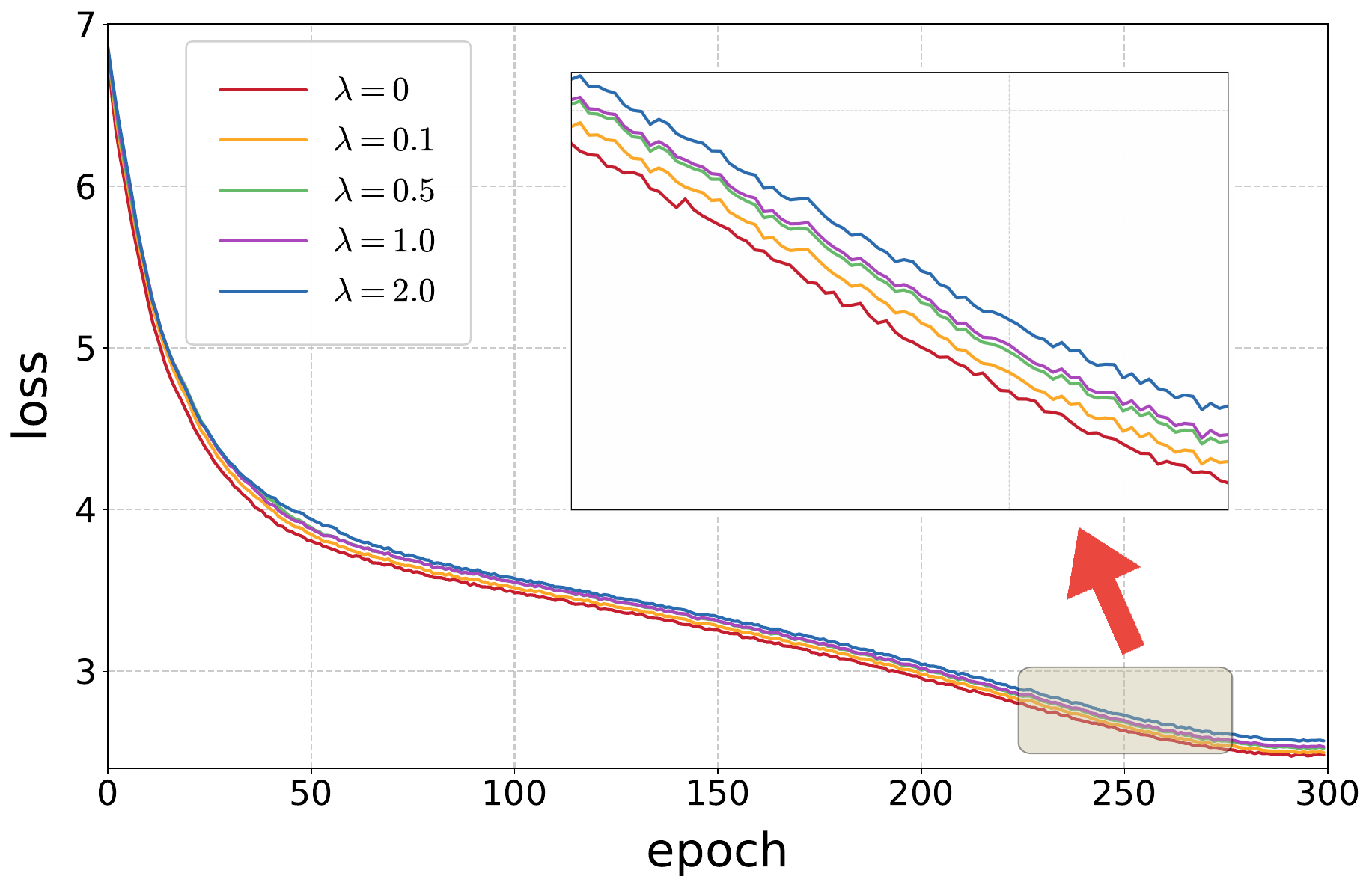}
    \vfill
    \caption{\textbf{Training loss curve for different center sensitivity (controlled by $\lambda$).}
    A larger $\lambda$ leads to higher training loss and poorer fitting ability.}
    \label{fig:center_sensitivity_training}
\end{minipage}
\hfill
\begin{minipage}[t]{0.54\linewidth}
    \centering
    \includegraphics[width=\linewidth]{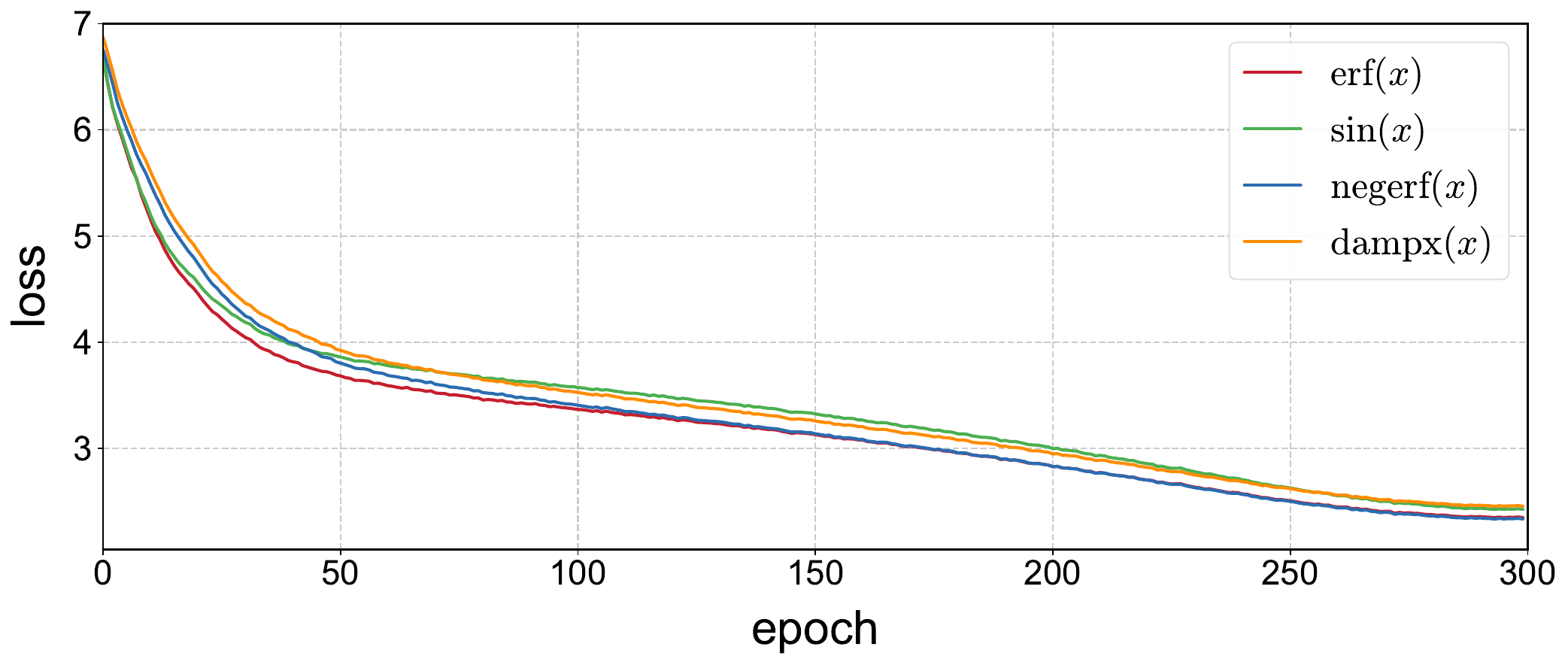}
    \vfill
    \caption{\textbf{Training loss curve for different monotonicity.}
    Monotonic functions consistently achieve lower training loss than non-monotonic functions.}
    \label{fig:monotonicity_train}
\end{minipage}

\end{figure}

We visualize the training losses obtained as $\lambda$ varies over \{0, 0.1, 0.5, 1.0, 2.0\} on the base point-wise function $\mathrm{erf}(x)$. As shown in \figref{fig:center_sensitivity_training}, training loss shows a clear monotonic trend: larger $\lambda$ consistently leads to higher loss, indicating that the width of the flat zone directly limits the model’s fitting capacity.

\vspace{-.3cm}
\subsection{Monotonicity}
We plot the training losses of four functions with distinct monotonicity patterns: the monotonically increasing \(\mathrm{erf}(x)\), the monotonically decreasing \(\mathrm{negerf}(x)\), the hump-shaped \(\mathrm{dampx}(x)\), and the oscillatory \(\sin(x)\). As shown in \figref{fig:monotonicity_train}, both increasing and decreasing monotonic functions achieve clearly lower training loss, indicating stronger fitting capacity. In contrast, the non-monotonic functions exhibit higher training loss. This behavior aligns closely with the top-1 accuracy trends observed on ImageNet-1K.

\vspace{-.3cm}
\section{Function Search Details}
\label{appendix:function_search_details}
In function search, a wide variety of common functional forms are systematically explored under the constraint of our function properties. The candidates range from polynomial and rational functions to the trigonometric and hyperbolic families, as well as various cumulative distribution functions. Beyond these common functional forms, we also experiment with their variants through translation, scaling, concatenation, and clipping. 

We categorize all candidate functions (see \tabref{tab:emprical_search_results}) into four groups: \textit{natural functions}, 
\textit{transformed basic functions}, \textit{clipped unbounded functions}, and 
\textit{canonical ratio functions}, and present detailed descriptions and 
visualizations of how each group is constructed.

\begin{figure}[htbp]
\centering

\begin{minipage}[b]{0.48\linewidth}
    \centering
    \includegraphics[width=0.8\linewidth]{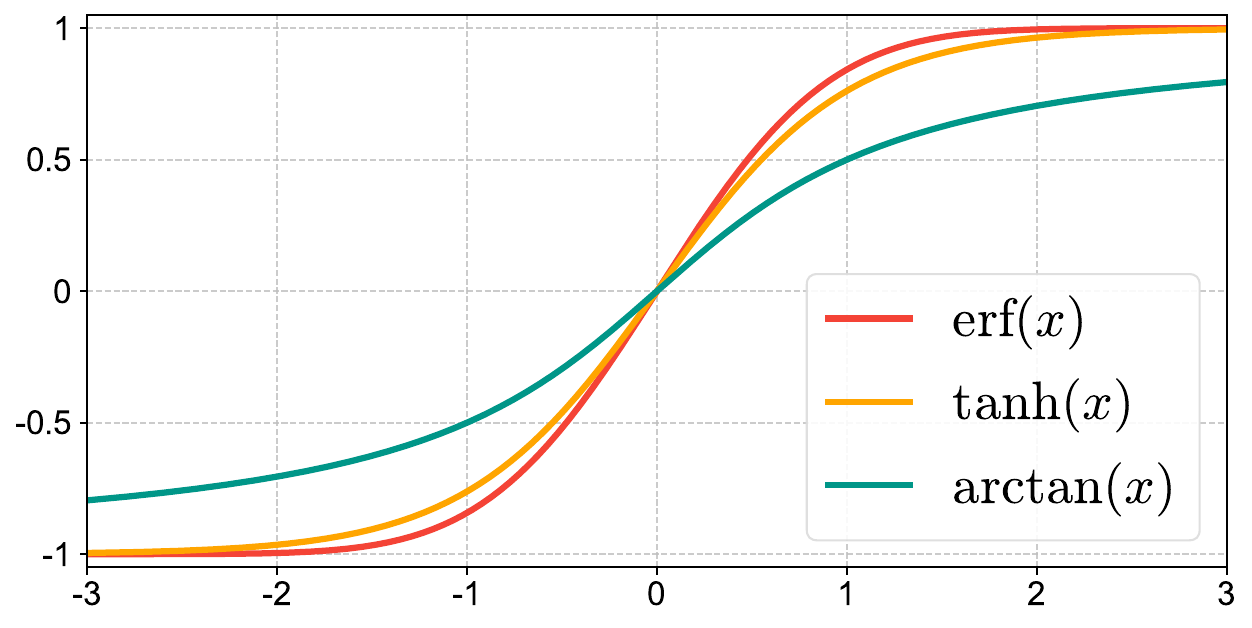}
    \captionof{figure}{\textbf{Visualization of natural functions.}}
    \label{fig:natural_function}

    \vspace{1em}

    \includegraphics[width=\linewidth]{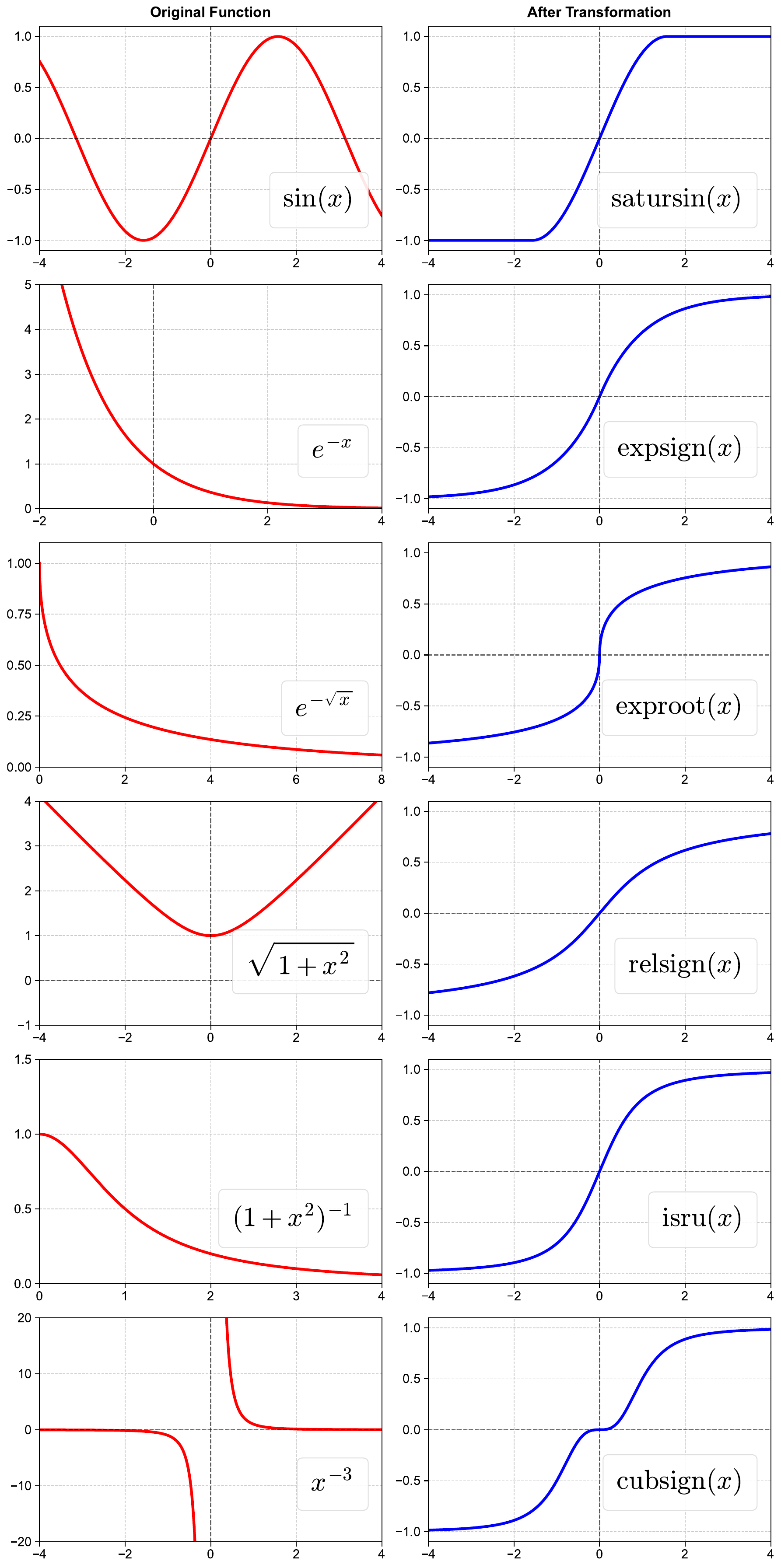}
    \captionof{figure}{\textbf{Visualization of transformed basic functions.}}
    \label{fig:transformed_basic_functions}
\end{minipage}
\hfill
\begin{minipage}[b]{0.48\linewidth}
    \centering
    \includegraphics[width=\linewidth]{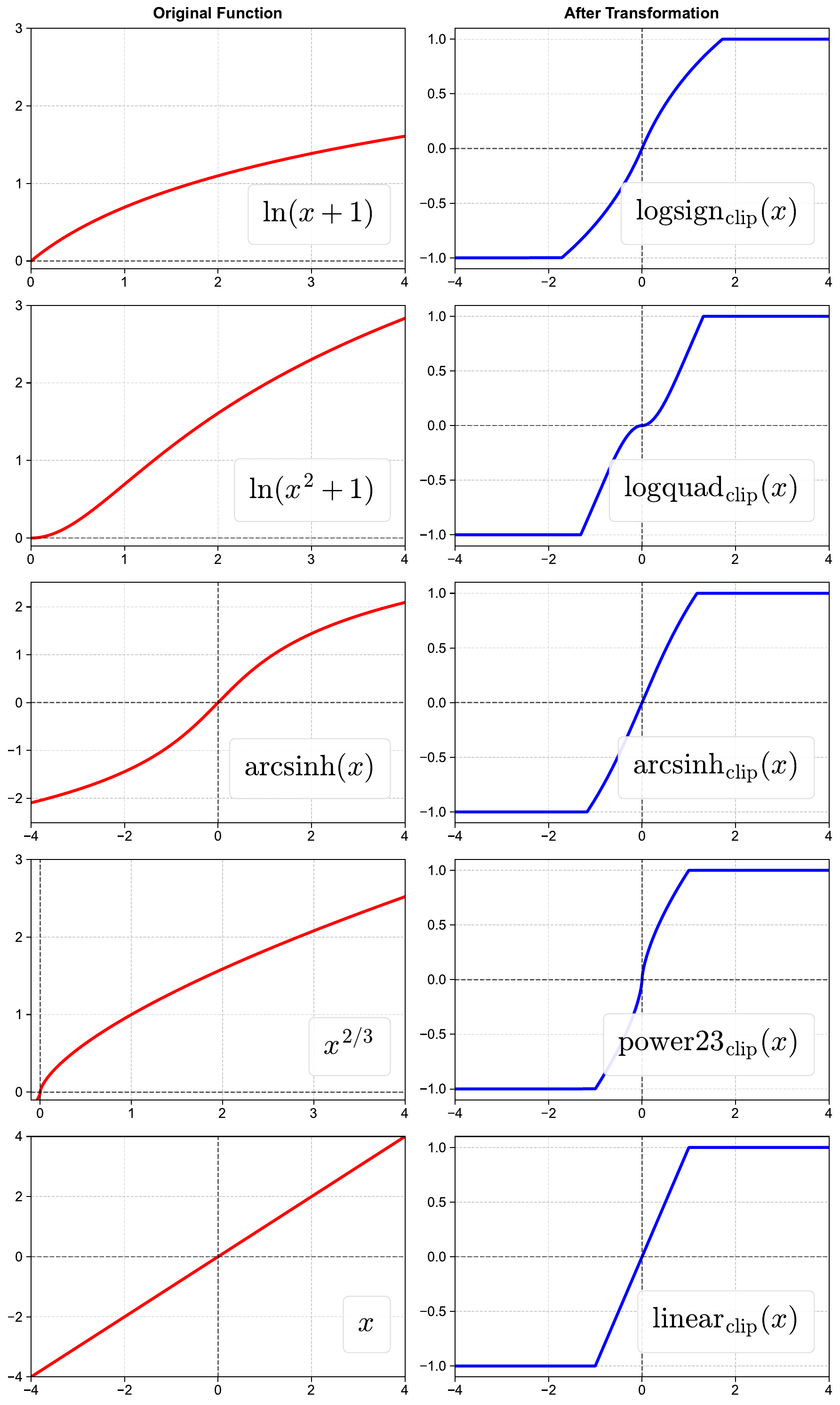}
    \captionof{figure}{\textbf{Visualization of clipped unbounded functions.}}
    \label{fig:clipped_unbounded_functions}

    \vspace{3em}

    \includegraphics[width=\linewidth]{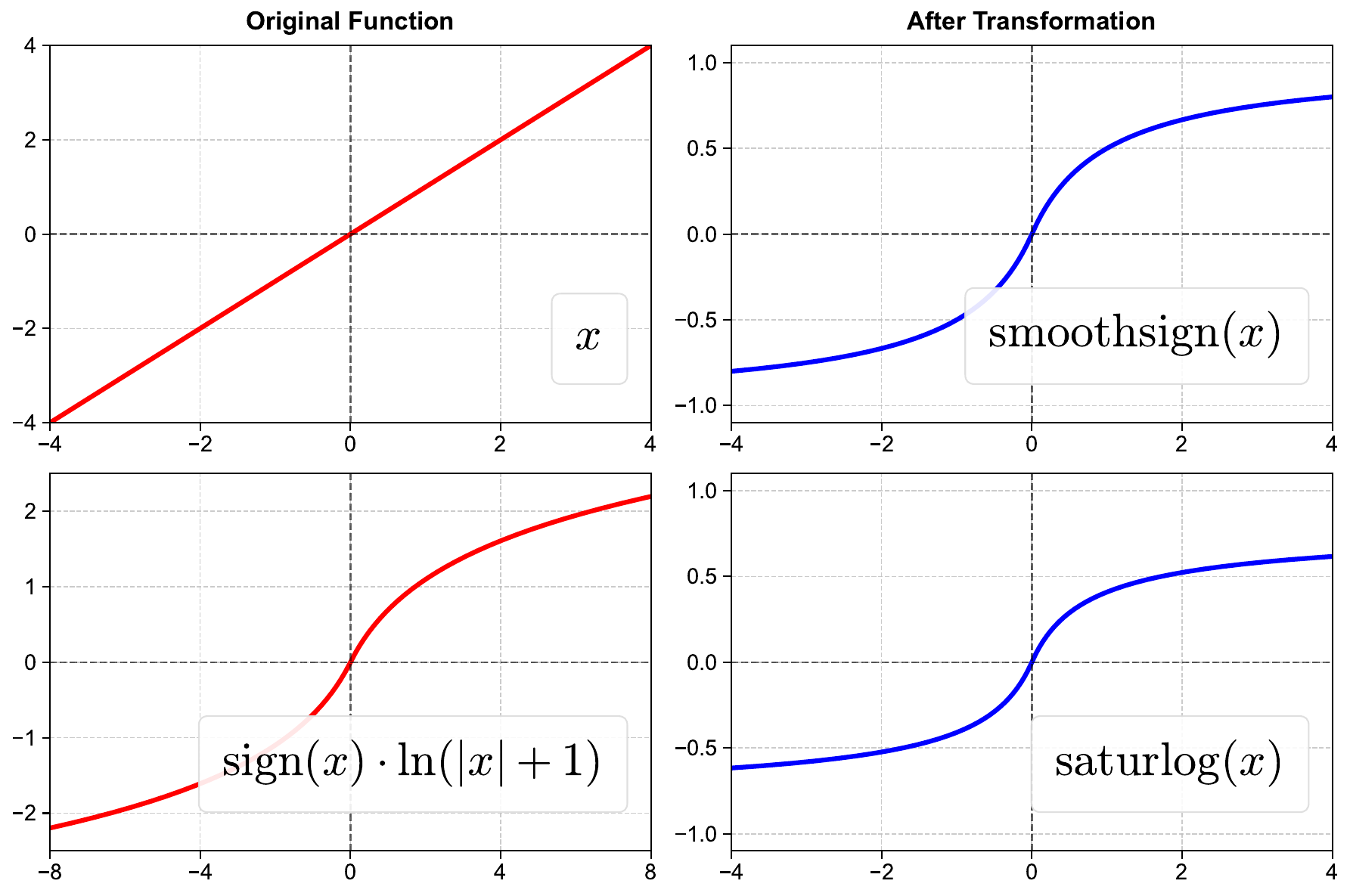}
    \captionof{figure}{\textbf{Visualization of canonical ratio functions.}}
    \label{fig:canonical_ratio_functions}
\end{minipage}

\end{figure}

\paragraph{Natural functions.} This category consists of three functions: $\mathrm{erf}(x)$, $\tanh(x)$, and $\arctan(x)$. As shown in \figref{fig:natural_function}, these functions naturally satisfy all the function properties, including \textit{zero-centeredness}, \textit{boundedness}, \textit{center sensitivity}, and \textit{monotonicity}. Among them, only $\arctan(x)$ is rescaled so that all three functions have their ranges unified to $[-1, 1]$.

\paragraph{Transformed basic functions.}
This category consists of six functions: $\mathrm{satursin}(x)$, $\mathrm{expsign}(x)$, $\mathrm{exproot}(x)$, $\mathrm{relsign}(x)$, $\mathrm{isru}(x)$, and $\mathrm{cubsign}(x)$. These functions are constructed by starting from simple and commonly used primitives, such as power functions and polynomial forms. Through transformations including translation, scaling, and rotation, we reshape their original structures so that they satisfy all four function properties while preserving the qualitative behavior of the underlying base functions, as shown in \figref{fig:transformed_basic_functions}.

\paragraph{Clipped unbounded functions.} 
This category consists of five functions: $\mathrm{logsign}(x)$, $\mathrm{logquad}(x)$, $\mathrm{arcsinh}(x)$, $\mathrm{power23}(x)$, and $\mathrm{linear}(x)$. These functions inherently satisfy \textit{zero-centeredness} and \textit{center sensitivity}. For $\mathrm{logsign_{clip}}(x)$, $\mathrm{logquad}(x)$, and $\mathrm{power23_{clip}}(x)$, either due to domain asymmetry or because the original form is not monotonic, we construct the negative branch by mirroring the positive side around the origin to ensure \textit{monotonicity}, as shown in \figref{fig:clipped_unbounded_functions}. To additionally enforce \textit{boundedness}, we clip their outputs to the interval $[-1, 1]$, which leads to improved performance in practice.

\paragraph{Canonical ratio functions.}
This category consists of two functions: $\mathrm{saturlog}(x)$ and $\mathrm{smoothsign}(x)$. 
Both functions are constructed using the canonical ratio form $\tfrac{f(x)}{|f(x)|+1}$, which 
naturally enforces \textit{boundedness} and \textit{monotonicity}. By selecting $f(x)$ to 
be an odd, zero-centered base function, the resulting ratio form automatically satisfies 
\textit{zero-centeredness} and \textit{center sensitivity} as well. As shown in \figref{fig:canonical_ratio_functions}, this construction yields smooth saturating behaviors that remain stable across a wide input range.

\section{Experimental Settings}
\label{appendix:experimental_settings}

\paragraph{Vision Transformers.} For all supervised classification experiments on ImageNet-1K, we adopt the training configurations summarized in \tabref{tab:supervised_vit}. ViT-B and ViT-L share the same hyperparameters, except that ViT-L employs a modified AdamW momentum setting with $(\beta_1{=}0.9,\, \beta_2{=}0.95)$ and a higher stochastic depth rate of $0.5$.

\begin{table}[h]
\tablestyle{6pt}{1.02}
\footnotesize
\begin{tabular}{c|c}
config & value \\
\shline
optimizer & AdamW \\
base learning rate & 4e-3 \\
weight decay & 0.05 \\
optimizer momentum & $\beta_1{=}0.9, \beta_2{=}0.999~(B), 0.95 ~(L)$ \\
effective batch size & 4096 \\
learning rate schedule & cosine decay \\
warmup epochs & 20 \\
training epochs & 300 \\
augmentation & rand-m9-mstd0.5-inc1 \\
label smoothing \citep{szegedy2016rethinking} & 0.1 \\
mixup \citep{zhang2017mixup} & 0.8 \\
cutmix \citep{yun2019cutmix} & 1.0 \\
random erase \citep{zhong2020random} & 0.25 \\
drop path \citep{huang2016deep} & 0.15 (B), 0.5 (L) \\
exp. moving average (EMA) & 0.9999
\end{tabular}
\caption{\textbf{Training Configurations of ViT.}}
\label{tab:supervised_vit}
\end{table}

\paragraph{Diffusion Transformers.} We use the official implementation \citep{peebles2023scalable} to train all DiT model sizes as shown in \tabref{tab:dit_training}. We observe that the default learning rate is suboptimal for the models in this work. For both the search function experiments and the final evaluation of Derf, we go through three learning rates, $1\times10^{-4}$, $2\times10^{-4}$, and $4\times10^{-4}$, for all models, whether they use LayerNorm or a point-wise function, and report the best result. We also observe that the zero initialization negatively affects the performance of Derf models and other point-wise function models. Therefore, we retain the zero initialization for LN models but remove it for the other models.

\begin{table}[h]
\tablestyle{6pt}{1.02}
\footnotesize
\begin{tabular}{c|c}
config & value \\
\shline
optimizer & AdamW \\
base learning rate & \{1e-4, 2e-4, 4e-4\} \\
weight decay & 0 \\
optimizer momentum & $\beta_1{=}0.9, \beta_2{=}0.999$ \\
effective batch size & 256 \\
learning rate schedule & constant \\
training epochs & 80 \\
exp. moving average (EMA) & 0.9999 \\
\end{tabular}
\caption{\textbf{Training Configurations of DiT.}}
\label{tab:dit_training}
\end{table}

\paragraph{Speech models.} For both wav2vec 2.0 models, we retain the first GroupNorm layer and the LayerNorm located after the convolutional feature extractor, since both primarily serve as data normalization to handle the unnormalized input data. We use the official implementation \citep{baevski2020wav2vec} for both the Base and Large models, keeping all hyperparameters identical to the original setup, as shown in \tabref{tab:wav2vec2_config}. The only change we make is running all models---whether normalization-based or point-wise-function-based---in \texttt{fp32} precision instead of the default \texttt{bf16}. We report the final validation loss.

\begin{table}[h]
\tablestyle{6pt}{1.02}
\footnotesize
\begin{tabular}{c|c}
config & value \\
\shline
optimizer & Adam \\
learning rate & 5e-4 (B), 3e-4 (L) \\
weight decay & 0.01 \\
optimizer momentum & $\beta_1{=}0.9, \beta_2{=}0.98$ \\
max tokens & 1400000 (B), 1200000 (L) \\
learning rate schedule & polynomial decay \\
warmup updates & 32000 (B), 20000 (L) \\
max updates & 400000 (B), 250000 (L) \\
dropout (input to encoder) & 0.1 \\
dropout (target features) & 0.1 \\
dropout (transformer) & 0.0 (B), 0.1 (L) \\
layer dropout & 0.05 (B), 0.2 (L) \\
feature grad mult & 0.1 \\
latent temp & [2,0.5,0.999995] (B), [2.0,0.1,0.999995] (L) \\
max sample size & 250000 (B), 320000 (L) \\
\end{tabular}
\caption{\textbf{Training Configurations of wav2vec 2.0.}}
\label{tab:wav2vec2_config}
\end{table}

\paragraph{DNA models.} For both the HyenaDNA model \citep{nguyen2023hyenadna} and the Caduceus model \citep{schiff2024caduceus}, we directly follow their official implementations without modifying hyperparameters, as shown in \tabref{tab:pretrain_configs}. In particular, Hyena uses LayerNorm and Caduceus uses RMSNorm. For our evaluation, we replace each model’s original normalization layer with Derf and report the average accuracy across all tasks.

\begin{table}[h]
\tablestyle{12pt}{1.02}
\footnotesize
\begin{tabular}{c|c}
config & value \\
\shline
optimizer & AdamW \\
learning rate & 6e-4 (H), 8e-3 (C) \\
sequence length & 1024 (H), 131072 (C) \\
effective batch size & 1024 (H), 8 (C) \\
training steps & 10000 (H), 50000 (C) \\
RC augmentation & true (H), false (C) \\
MLM probability & 0.0 (H), 0.15 (C) \\
bidirectional & false (H), true (C) \\
\end{tabular}
\caption{\textbf{Training Configurations of HyenaDNA and Caduceus.} H denotes HyenaDNA, C denotes Caduceus.}
\label{tab:pretrain_configs}
\end{table}

\paragraph{Language models.} For the GPT-2 (124M) model, we follow the hyperparameters as shown in \tabref{tab:gpt2_training}. For Derf and DyT, we configure the $\alpha$ initialization separately for the point-wise function layer following the attention layer and for the other point-wise function layers. We try multiple combinations of these initialization settings and report the best validation loss.

\begin{table}[h]
\tablestyle{6pt}{1.02}
\footnotesize
\begin{tabular}{c|c}
config & value \\
\shline
optimizer & AdamW \\
base learning rate & 6e-4 \\
$\alpha$ initialization attention & \{0.5, 1.0, 2.0, 4.0\}\\
$\alpha$ initialization other & \{0.1, 0.3, 0.5, 1.0\} \\
weight decay & 0.1 \\
optimizer momentum & $\beta_1{=}0.9, \beta_2{=}0.95$ \\
gradient clipping & 1.0 \\
block size & 1024 \\
gradient accumulation steps & 40  \\
effective batch size & 491,520  \\
learning rate schedule & cosine decay \\
warmup iterations & 2,000 \\
training iterations & 300,000 \\
dropout & 0.0 \\
mixed precision & \texttt{bf16} \\
\end{tabular}
\caption{\textbf{Training Configurations of GPT-2 (124M).}}
\label{tab:gpt2_training}
\end{table}

\section{Additional Results}
\label{appendix:additional_results}
Beyond evaluating each model with its default normalization layer (typically LN), we additionally test RMSNorm and GroupNorm (GN) to enable a more complete comparison. RMSNorm is widely used in modern large language models, including T5 \citep{raffel2020exploring}, LLaMA \citep{touvron2023llama, touvron2023llama2, dubey2024llama}, Qwen \citep{bai2023qwen, yang2024qwen2}, and DeepSeek \citep{liu2024deepseek, guo2025deepseek}, while GN is employed in several vision architectures, including ConvNeXt \citep{liu2022convnet}, DETR \citep{carion2020end}, and Swin Transformer \citep{liu2021swin}.

All evaluations follow the same experimental settings described in the previous section. These additional results show that Derf not only surpasses the default choices used in each model, but also outperforms the other normalization alternatives we evaluate.

\paragraph{Vision Transformers.} For both ViT-Base and ViT-Large \citep{dosovitskiy2020image}, the default normalization layer is LayerNorm. To complement the results, we also evaluate RMSNorm \citep{zhang2019root} and GN \citep{wu2018group} as additional replacements in \tabref{tab:addtional_vit_results}. Compared to all other methods, \methodname{} achieves clearly higher top-1 accuracy, demonstrating its effectiveness in vision transformer architectures.

\begin{table}[h!]
\vspace{0.3em}
\centering
\tablestyle{7.5pt}{1.15}
\begin{tabular}{lccccc}
\toprule
model & LN & DyT & \methodname{} & RMSNorm & GN \\
\midrule
ViT-B & 82.3\% & 82.5\% & \textbf{82.8}\% & 82.4\% & 82.5\%\\
ViT-L & 83.1\% & 83.6\% & \textbf{83.8}\% & 83.0\% & 83.1\%\\
\midrule
\end{tabular}
\vspace{-0.3em}
\caption{\textbf{Supervised classification accuracy on ImageNet-1K.} \methodname{} achieves higher top-1 accuracy than all other methods on different model sizes.}
\label{tab:addtional_vit_results}
\end{table}

\paragraph{Diffusion Transformers.} For DiT models \citep{peebles2023scalable}, we additionally evaluate RMSNorm \citep{zhang2019root} as an alternative normalization layer and compare its performance with LN, DyT, and \methodname{}. As shown in \tabref{tab:additional_dit_results}, \methodname{} achieves a clear improvement in FID compared to all other methods.

\begin{table}[h!]
\vspace{0.3em}
\centering
\tablestyle{11.5pt}{1.15}
\begin{tabular}{lccccc}
\toprule
model & LN & DyT & \methodname{} & RMSNorm \\
\midrule
DiT-B/4 & 64.93 & 63.94 & \textbf{63.23} & 65.08 \\
DiT-L/4 & 45.91 & 45.66 & \textbf{43.94} &  45.02 \\
DiT-XL/2 & 19.94 & 20.83 & \textbf{18.92} & 20.76 \\
\midrule
\end{tabular}
\vspace{-0.3em}
\caption{\textbf{Image generation quality (FID) on ImageNet.} Lower FID indicates better image generation quality. \methodname{} achieves lower FID scores than all other methods across different DiT models.}
\label{tab:additional_dit_results}
\end{table}

\paragraph{Speech models.}
For two wav2vec 2.0 Transformer models \citep{baevski2020wav2vec}, we additionally evaluate RMSNorm \citep{zhang2019root} as an alternative normalization layer and compare its performance with LN, DyT, and \methodname{} in \tabref{tab:additional_wav2vec_results}. Compared to other methods, \methodname{} yields lower validation loss on different model sizes

\begin{table}[h!]
\vspace{0.3em}
\centering
\tablestyle{9pt}{1.15}
\begin{tabular}{lccccc}
\toprule
model & LN & DyT & \methodname{} & RMSNorm \\
\midrule
wav2vec 2.0 Base & 1.95 & 1.95 & \textbf{1.93} & 1.95 \\
wav2vec 2.0 Large & 1.92 & 1.91 & \textbf{1.90} & 1.93 \\
\midrule
\end{tabular}
\vspace{-0.3em}
\caption{\textbf{Speech pretraining validation loss on the LibriSpeech dataset.} \methodname{} achieves lower validation loss than all other methods across two wav2vec~2.0 models.}

\label{tab:additional_wav2vec_results}
\end{table}

\paragraph{DNA models.} For the HyenaDNA model \citep{nguyen2023hyenadna} and the Caduceus model \citep{schiff2024caduceus}, we additionally evaluate both LayerNorm and RMSNorm for each architecture, regardless of their default choices, and compare their performance with DyT and \methodname{} in \tabref{tab:additional_dna_results}.

\begin{table}[h!]
\vspace{0.3em}
\centering
\tablestyle{10.5pt}{1.15}
\begin{tabular}{lccccc}
\toprule
model & LN & DyT & \methodname{} & RMSNorm \\
\midrule
Hyena & 85.2\% & 85.2\% & \textbf{85.7}\% & 85.2\% \\
Caduceus & 87.0\%  & 86.9\% & \textbf{87.3}\% & 86.9\% \\
\midrule
\end{tabular}
\vspace{-0.3em}
\caption{\textbf{DNA classification accuracy on the GenomicBenchmarks dataset}, averaged over each subtask. \methodname{} consistently outperforms other methods across two different DNA models.} 
\label{tab:additional_dna_results}
\vspace{-0.3cm}
\end{table}

\paragraph{Language models.} For the GPT-2 (124M) model, we additionally evaluate RMSNorm \citep{zhang2019root} for a more complete comparison of normalization choices. As shown in \tabref{tab:additional_llm_results}, \methodname{} achieves comparable performance to both LN and RMSNorm, while clearly outperforming DyT.

\begin{table}[h!]
\vspace{0.3em}
\centering
\tablestyle{13pt}{1.15}
\begin{tabular}{lccccc}
\toprule
model & LN & DyT & \methodname{} & RMSNorm \\
\midrule
GPT-2  & \textbf{2.94} & 2.97 & \textbf{2.94} & 2.95 \\
\midrule
\end{tabular}
\vspace{-0.3em}
\caption{\textbf{GPT-2 validation loss on the OpenWebText dataset.} \methodname{} matches the performances of both LayerNorm and RMSNorm while achieving lower validation loss than DyT.}
\label{tab:additional_llm_results}
\vspace{-0.3cm}
\end{table}

\vspace{-0.3cm}
\section{Loss Calculation Details}
\label{appendix:loss_calculation_details}
\vspace{-0.1cm}

\paragraph{Vision Transformers.}
For ViT models, we measure fitting capacity under a deterministic evaluation setup. We switch the model to evaluation mode, disable drop-path, mixup, cutmix, label smoothing, and all data augmentations, and apply only the standard test-time preprocessing (center crop and normalize). The cross-entropy loss is then computed on the training set and averaged over all samples.
\vspace{-0.1cm}

\paragraph{Diffusion Transformers.} For DiT models, we evaluate fitting capacity by switching the model to evaluation mode. We apply the standard test-time preprocessing (center crop, random horizontal flip, and normalize). Since DiT does not employ drop-path, no stochastic regularization needs to be disabled. We then compute the diffusion MSE loss over the first 100 training batches and report the average.
\vspace{-0.1cm}

\paragraph{Other models.} For all other models, wav2vec 2.0, HyenaDNA, Caduceus, and GPT2, we simply apply the same procedure: use the standard test-time preprocessing, disable drop-path or dropout when present, and compute the training loss over the full training set, reporting the average.

\vspace{-0.3cm}
\section{Computation Cost}
\label{appendix:computation_cost}
\begin{figure}[h!]
\centering

\begin{minipage}[b]{0.64\linewidth}
  \centering
  \includegraphics[width=.8\linewidth]{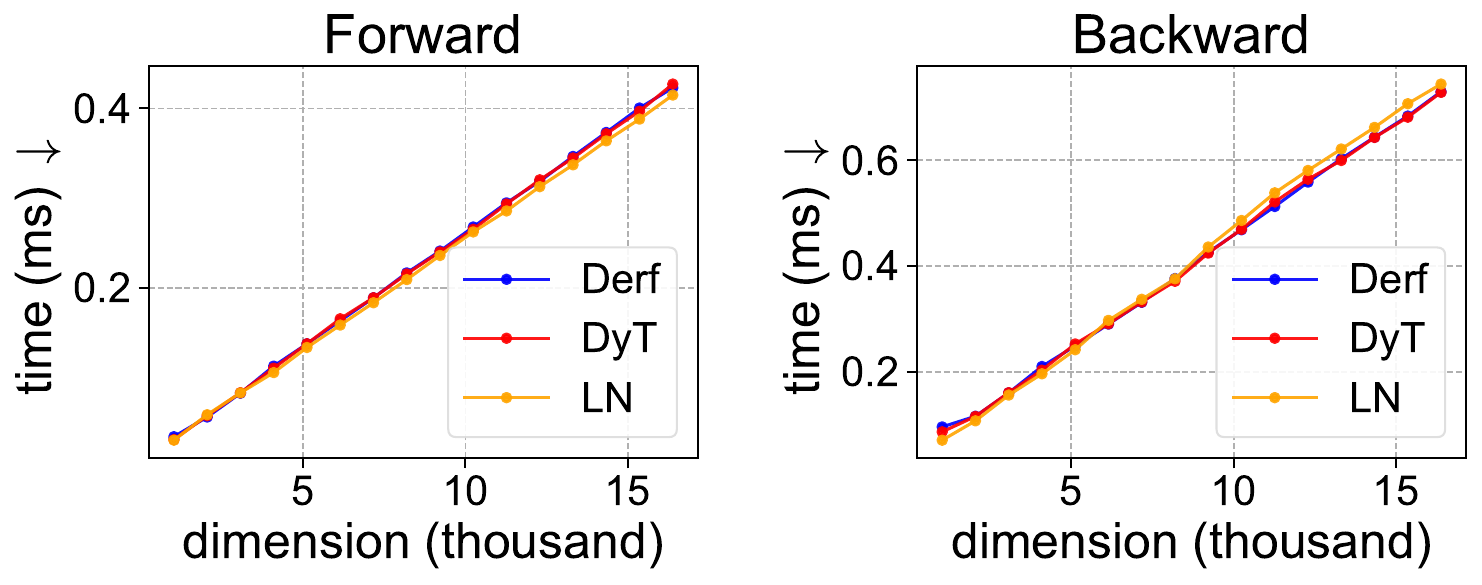}
  \caption{Forward and backward wall-clock runtime comparison for DyT, Derf, and LayerNorm.}
  \label{fig:efficiency}
\end{minipage}\hfill
\begin{minipage}[b]{0.34\linewidth}
  \centering
  \tablestyle{5.5pt}{1.35}
  \begin{tabular}{@{\hspace{4pt}}lcc@{\hspace{4pt}}}
    \toprule
      & Forward & Backward \\
    \midrule
    Derf  & 0.42 & 0.73 \\
    DyT & 0.42 & 0.73 \\
    LN & 0.42 & 0.74 \\
    \bottomrule
  \end{tabular}
  \vspace{0.6cm}
  \captionof{table}{Time (ms) for running 15k hidden dimension.}
  \label{tab:efficiency}
\end{minipage}

\end{figure}

We implement DyT, Derf, and LayerNorm using custom Triton kernels and measured forward and backward runtimes of a single norm layer operation under a unified setup. All operators use comparable fusion levels, memory access patterns, and precision, and are benchmarked using the same batch size and sequence length, across a range of hidden dimensions. As shown in \figref{fig:efficiency} and \tabref{tab:efficiency}, Derf and DyT exhibit nearly identical runtime in both forward and backward passes across all practical hidden dimensions. In the forward pass, Derf and LayerNorm have comparable runtime across dimensions. In the backward pass, Derf can be slightly slower than LayerNorm at small hidden dimensions, but becomes consistently faster as the dimension increases, which aligns with the growing reduction and synchronization overhead in norm layers.

\newpage

\end{document}